\documentclass[letterpaper, 10 pt, conference]{ieeeconf}  % Comment this line out if you need a4paper

\IEEEoverridecommandlockouts                              % This command is only needed if 
\usepackage{amsmath,amsfonts}
\usepackage{amssymb}
\usepackage{graphicx}
\usepackage{algorithm}
\usepackage{array}
\usepackage[caption=false,font=normalsize,labelfont=sf,textfont=sf]{subfig}
\usepackage{textcomp}
\usepackage{stfloats}
\usepackage{url}
\usepackage{verbatim}
\usepackage{graphicx}
\usepackage{cite}
\usepackage{multirow}
\usepackage{multicol}
\usepackage{colortbl}
\usepackage[table]{xcolor}
\usepackage{threeparttable}
\usepackage{pifont}
\usepackage{circledsteps}
\usepackage{caption}
\usepackage{algpseudocode}

\algrenewcommand\textproc{}

\title{\LARGE \bf
Depth-Aware Range Image-Based Model for Point Cloud Segmentation
}

\author{Bike Chen, Antti Tikanmäki, Juha Röning% <-this % stops a space
}

\begin{document}

\maketitle
\thispagestyle{empty}
\pagestyle{empty}

%%%%%%%%%%%%%%%%%%%%%%%%%%%%%%%%%%%%%%%%%%%%%%%%%%%%%%%%%%%%%%%%%%%%%%%%%%%%%%%%
\begin{abstract}

\textbf{Point cloud segmentation (PCS) aims to separate points into different and meaningful groups. The task plays an important role in robotics because PCS enables robots to understand their physical environments directly. To process sparse and large-scale outdoor point clouds in real time, range image-based models are commonly adopted. However, in a range image, the lack of explicit depth information inevitably causes some separate objects in 3D space to touch each other, bringing difficulty for the range image-based models in correctly segmenting the objects. Moreover, previous PCS models are usually derived from the existing color image-based models and unable to make full use of the implicit but ordered depth information inherent in the range image, thereby achieving inferior performance. In this paper, we propose \textbf{D}epth-\textbf{A}ware \textbf{M}odule (DAM) and Fast FMVNet V3. DAM perceives the ordered depth information in the range image by explicitly modelling the interdependence among channels. Fast FMVNet V3 incorporates DAM by integrating it into the last block in each architecture stage. Extensive experiments conducted on SemanticKITTI, nuScenes, and SemanticPOSS demonstrate that DAM brings a significant improvement for Fast FMVNet V3 with negligible computational cost.}
\end{abstract}

%%%%%%%%%%%%%%%%%%%%%%%%%%%%%%%%%%%%%%%%%%%%%%%%%%%%%%%%%%%%%%%%%%%%%%%%%%%%%%%%
\section{INTRODUCTION}
% Step 1: Background.
Point cloud segmentation (PCS) is to classify each point in point clouds. The task plays an essential role in robots and autonomous driving vehicles. Specifically, the pointwise predictions enable robots to directly understand the 3D physical environments~\cite{semantickitti_2019_behley,nuscenes_panoptic,semanticposs_2020}, such as searching for free space and detecting objects for obstacle avoidance. Besides, the per-point predictions can be used to build a semantic map~\cite{suma++_2019,sa_loam_2021} for robot navigation. 

% Step 2: What are the assumptions?
This paper focuses on the range image-based PCS models on the large-scale outdoor point clouds~\cite{semantickitti_2019_behley,nuscenes_panoptic,semanticposs_2020}. The range image is a compact representation of a sparse and large-scale point cloud. Corresponding range image-based models, such as FIDNet~\cite{fidnet_2021} and Fast FMVNet V2~\cite{pdm2024}, can predict the point cloud at high speed with competitive segmentation accuracy when compared with voxel- and point-based models, such as MinkUNet~\cite{minkowski2019}, SPVCNN~\cite{spvnas_2020}, WaffleIron~\cite{waffleiron23}, and PTv3~\cite{pointtransv32024}.

\begin{figure}[t]
	\centering
	\includegraphics[width=1.0\columnwidth]{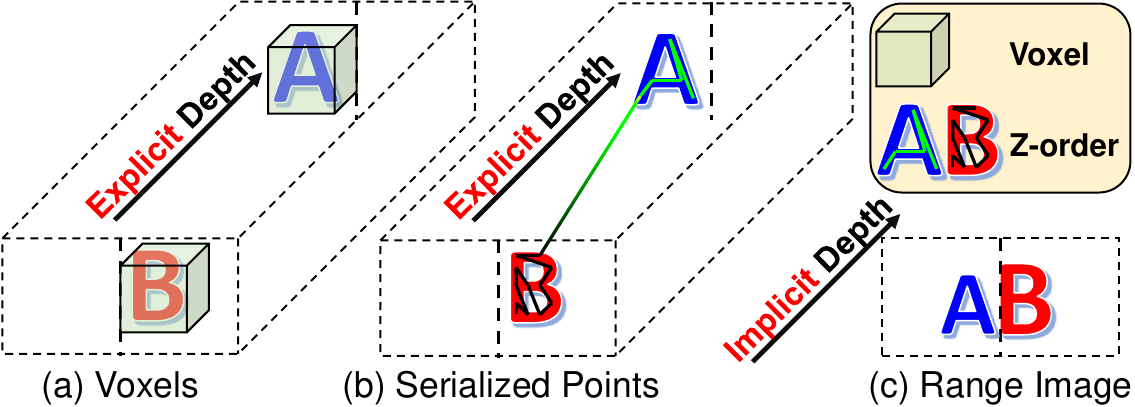}
	\caption{(a) In a point cloud, objects \textcolor{blue}{A} and \textcolor{red}{B} remain separate along the depth direction after all points are voxelized~\cite{minkowski2019,spvnas_2020}. (b) After the points are serialized by space-filling curves such as Z-order~\cite{pointtransv32024}, the objects \textcolor{blue}{A} and \textcolor{red}{B} are also separate naturally in 3D space. We consider that the voxel and point representations have explicit depth information. (c) However, the oject \textcolor{blue}{A} touches the object \textcolor{red}{B} in the range image although they are far apart in 3D space. We think that the range image contains implicit depth information.}
	\label{fig:voxel_points_range_image}  
\end{figure}

% Step 3: What is the motivation of your work? What is the significance of your work?
However, unlike voxel and point representations with explicit depth information, the range image representation only has implicit depth information and inevitably brings difficulty in accurately making pointwise predictions for the existing range image-based models. In the voxel- or point-based point cloud, some objects are naturally far apart along the depth direction in 3D space. For example, in Figs.~\ref{fig:voxel_points_range_image}(a) and (b), the object \textcolor{blue}{A} is distant from the object \textcolor{red}{B}. Hence, we consider that the voxel and point representations have explicit depth information. By contrast, in Fig.~\ref{fig:voxel_points_range_image}(c), the separate objects \textcolor{blue}{A} and \textcolor{red}{B} touch each other in the range image. Therefore, the range image representation only contains implicit depth information. Also, the ``compressed" range image cannot directly and accurately describe the 3D shapes of objects, posing a challenge to the performance improvement of the existing range image-based models.  

\begin{figure}[t]
	\centering
	\includegraphics[width=0.95\columnwidth]{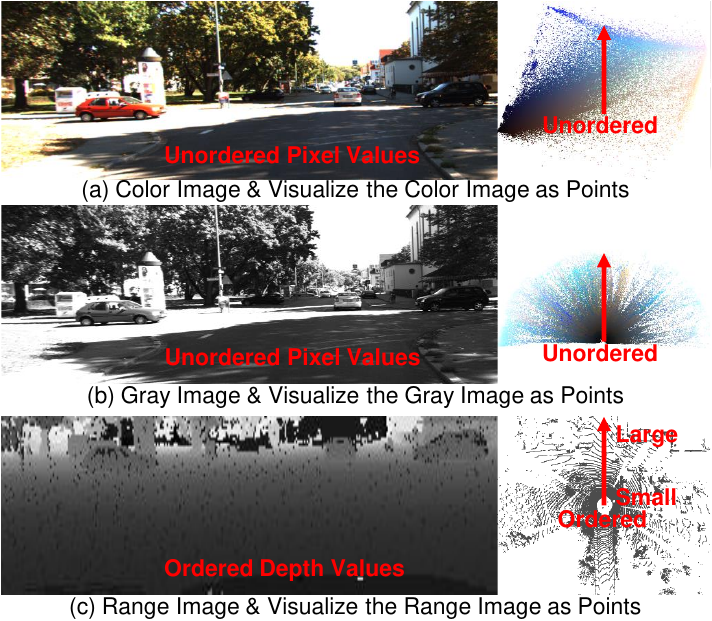}
	\caption{(a) A color image~\cite{kitti12} and the pixels visualized in 3D space where RGB values serve as the coordinates and colors. (b) A gray image and the pixels visualized in 3D space where the pixel values are transformed to the coordinates by a spherical model and the above RGB values act as the colors. In the color and gray images, the pixel value changes disorderly. Also, we cannot find any meaningful objects in the above ``pixel clouds". (c) The range image and corresponding point cloud. The depth value in the range image varies with the distance of the target object from the LiDAR sensor. The objects in the point cloud are meaningful. We think that the depth values are ordered.}
	\label{fig:color_gray_range_image}  
\end{figure}

% Step 4: Why can others not do this?
More importantly, the previous range image-based models, such as FIDNet~\cite{fidnet_2021} and Fast FMVNet V2~\cite{pdm2024}, are unable to take full advantage of the implicit but ordered depth information inherent in the range image because they are derived from color image-based models like ResNet~\cite{resnet_2016} and ConvNeXt~\cite{convnext2022}. Here, we argue that depth values in the range image are ordered but pixel values in the color image are unordered. For example, in Fig~\ref{fig:color_gray_range_image}(c), we see that the depth value changes in an orderly way, from small to large with the increased distance of the objects from the light detection and ranging (LiDAR) sensor. By contrast, the pixel value changes disorderly in the color or gray image (see Figs.~\ref{fig:color_gray_range_image}(a) and (b)). Hence, the depth values are ordered to form the objects in the range image and different from the unordered pixel values, indicating that the color image-based models are not suitable to use in directly processing the range image. Additionally, although we can adopt some existing channel attention modules from the color image-based models, such as the SE (squeeze-and-excitation) block~\cite{senet18} and GRN (global response normalization)~\cite{convnextv223}, to enable the range image-based PCS models to perceive the implicit depth information, these modules still fail to consider the order of the depth values. Correspondingly, the models still obtain inferior performance. Therefore, a depth-aware module targeting the existing range image-based models is required.

% Step 5: Therefore, you propose a method which can solve previous problems (Why).
To make full use of the implicit but ordered depth information inherent in the range image, we design a \textbf{D}epth-\textbf{A}ware \textbf{M}odule (DAM) and introduce Fast FMVNet V3. DAM utilizes the depth information by dynamically weighting the channel-wise feature responses. DAM contains two key parts: (1) a global average pooling (GAP), which effectively summarizes the global contextual information in each feature map, and (2) a sinusoidal positional encoding (SPE), which considers the order of the depth values inherent in the range image. Fast FMVNet V3 is the upgraded Fast FMVNet V2~\cite{pdm2024}. In Fast FMVNet V3, we integrate DAM into the last block in each stage of the backbone.

% Step 6: Briefly describe the experimental results and claim you contribution. 
Extensive experiments are conducted on the popular SemanticKITTI~\cite{semantickitti_2019_behley}, nuScenes~\cite{nuscenes_panoptic}, and SemanticPOSS~\cite{semanticposs_2020} datasets. On the SemanticKITTI test dataset, without any test-time augmentation and ensemble tricks, our Fast FMVNet V3 achieves the 69.6\% mean intersection-over-union (mIoU) score with 25.5 frames per second (FPS). On the nuScenes and SemanticPOSS datasets, Fast FMVNet V3 reaches 76.6\% and 56.1\% mIoU scores, respectively. The experimental results demonstrate the superiority of the proposed DAM and Fast FMVNet V3. The contributions in this paper are summarized as follows:

\begin{itemize}
	\item A \textbf{D}epth-\textbf{A}ware \textbf{M}odule (DAM) is designed to make full use of the implicit but ordered depth information in the range image by dynamically modelling the interdependence among the channels. 
	
	\item Fast FMVNet V3 is introduced, which incorporates DAM into the last block in each architecture stage. Fast FMVNet V3 can achieve new state-of-the-art performance regarding the speed-accuracy trade-off on the SemanticKITTI, nuScenes, and SemanticPOSS datasets.
\end{itemize}

\section{RELATED WORK}
In this section, we briefly review the related range image-based point cloud segmentation models, attention modules, and position encoding techniques. 

\subsection{Range Image-Based Models}
Range image-based models can run at high speed and achieve competitive performance when compared with voxel-based models~\cite{minkowski2019,spvnas_2020}. Existing range image-based models are usually derived from the image classification and semantic image segmentation models because of the similarity between the range image and the color image. For example, SqueezeSeg~\cite{squeezeseg} uses the revised SqueezeNet~\cite{squeezenet_2016} as the backbone. RangeNet++~\cite{rangenet++} utilizes the modified DarkNet~\cite{yolov3_2018} to make per-point predictions. CENet~\cite{cenet_2022} and FIDNet~\cite{fidnet_2021} are built on the ResNet~\cite{resnet_2016}. RangeViT~\cite{rangevit_2023} is derived from ViT~\cite{vit_iclr_2021}. Fast FMVNet~\cite{filling_missing2024} and Fast FMVNet V2~\cite{pdm2024} opt the modified ConvNeXt-Tiny~\cite{convnext2022} as their backbones. In addition, to compensate for the loss of information in the range image, various post-processing components are applied, such as CRF (conditional random field)~\cite{squeezeseg}, $K$NN ($K$-nearest neighbor search)~\cite{rangenet++}, NLA (nearest label assignment)~\cite{fidnet_2021}, and PDM (pointwise decoder module)~\cite{pdm2024}. However, the existing range image-based models fail to consider the implicit but ordered depth information inherent in the range image. In this paper, we extend Fast FMVNet V2 and introduce Fast FMVNet V3 by incorporating a novel depth-aware module (DAM) to fully utilize the depth information. Fast FMVNet V3 can significantly surpass its counterpart in segmentation accuracy.

\subsection{Attention Modules}
The proposed depth-aware module (DAM) can be seen as a channel attention component, so we briefly review the associated attention modules here. The attention components can be roughly grouped into spatial attention and channel attention modules. For the spatial attention components, the works SegNeXt~\cite{segnext22}, MetaFormer~\cite{metaformer22}, and FocalNet~\cite{focalnet22} introduce MSCA (multi-scale convolutional attention), a pooling operation, and a focal modulation module to completely replace the self-attention mechanism in the transformer architectures, respectively. These attention modules improve the model performance in color image-based tasks such as image classification. However, we empirically find that these attention components cannot work well for the models trained on the range images, probably because of the ordered depth values. Besides, in the 3D object detection field, RangeDet~\cite{rangedet21} introduces the MKC (meta-kernel convolution) to dynamically extract geometric features in the range image. However, MKC requires high computational cost due to the unfolding operation. For the channel attention modules, the works SENet~\cite{senet18}, CBAM~\cite{cbam18}, and ConvNeXt V2~\cite{convnextv223} introduce the SE (squeeze-and-excitation) block, channel attention module, and GRN (global response normalization) to model the interdependent relationships among the channels, respectively. However, these channel attention modules fail to consider the order of the depth values in the range image. Instead, this paper introduces a depth-aware module that considers the ordered depth information in the range image.

\subsection{Positional Encoding}
The introduced depth-aware module (DAM) contains a positional encoding (PE) part, so we briefly review related PE techniques here. Different from convolution-based networks, which can implicitly learn positions in the 2D image, transformer-based networks require PE~\cite{attention2017,impactpe23} to understand the order of the sequential data, especially in the natural language processing field. The commonly used PE techniques are sinusoidal positional encoding~\cite{attention2017}, T5's relative bias, rotary positional embedding~\cite{roformer24,llama23}, and ALiBi positional embedding~\cite{bloom23}. All these PE techniques are employed with transformers. However, this paper adopts the convolution-based PCS model, which does not have $Q$ and $K$ vectors. Hence, we choose the typical sinusoidal positional encoding (SPE) without binding with the $Q$ and $K$ vectors. SPE serves as the sub-module of the proposed DAM to enable Fast FMVNet V3 to understand the order of the depth values in the range image.

\section{MODEL}
In this section, we first detail the proposed depth-aware module (DAM) and then describe the architecture of the introduced Fast FMVNet V3.

\begin{figure*}[t]
	\centering
	\includegraphics[width=1.39\columnwidth]{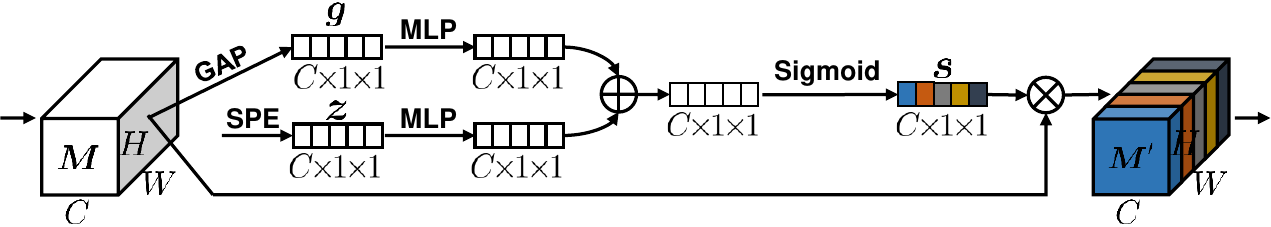}
	\caption{Overview of the proposed depth-aware module (DAM). First, feature maps $\boldsymbol{M}$ go through a global average pooling (GAP) layer to output the vector $\boldsymbol{g}$. The sinusoidal positional encoding (SPE) generates the vector $\boldsymbol{z}$. Then, both $\boldsymbol{g}$ and $\boldsymbol{z}$ pass through a shared multi-layer perceptron (MLP). The outputs from $\boldsymbol{g}$ and $\boldsymbol{z}$ are summed and go through a Sigmoid function to produce a scale $\boldsymbol{s}$. Finally, the scale $\boldsymbol{s}$ is multiplied by the feature maps $\boldsymbol{M}$ to make the depth-aware feature maps $\boldsymbol{M}^{\prime}$.}
	\label{fig:depth_aware_module}  
\end{figure*}

\subsection{Depth-Aware Module}
The depth-aware module (DAM) aims to make full use of the ordered depth information in the range image. Here, we first describe how to produce the positions to explicitly encode the depth information in the range image. Then, we detail the proposed DAM.

\subsubsection{Sinusoidal Positional Encoding}
Inspired by the natural language processing (NLP) work~\cite{attention2017}, we here utilize the sinusoidal position encoding (SPE) to indicate the depth information. SPE does not depend on the $Q$ and $K$ vectors of the transformers. It can run at high speed. Also, it can be easily integrated into DAM. SPE is expressed by the following Eq.~(\ref{eq:positional_encoding}),
\begin{equation}
	\text{SPE}(pos, d) =
	\begin{cases} 
		\text{sin}(\frac{pos}{10000^{d/d_{\text{model}}}}), & d \bmod 2 = 0 \\
		\text{cos}(\frac{pos}{10000^{d/d_{\text{model}}}}), & d \bmod 2 = 1
	\end{cases},
	\label{eq:positional_encoding}
\end{equation}
where $pos$ means the positions; $d$ indicates the dimensions; $d_{\text{model}}$ shows the model dimension and is equal to the model channels $C$.

\begin{figure}[t]
	\centering
	\includegraphics[width=0.98\columnwidth]{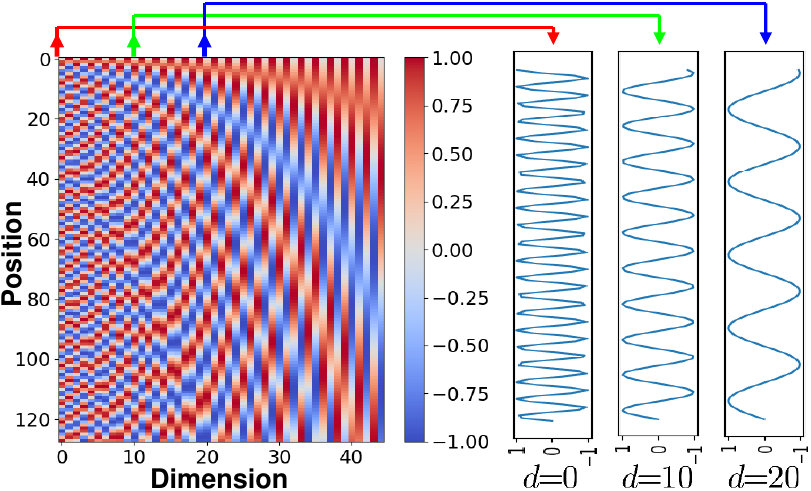}
	\caption{Left: visualization of sinusoidal position encoding~\cite{attention2017}; Right: sinusoids formed by three groups of position values obtained under the dimensions of 0, 10, and 20.}
	\label{fig:position_encoding}  
\end{figure}

However, different from the position encoding in the NLP work~\cite{attention2017}, SPE here is a vector $\boldsymbol{z} \in \mathbb{R}^C$ applied for the channels. The position values can be obtained by fixing one dimension. For example, Fig.~\ref{fig:position_encoding} shows that three groups of positions are obtained under the dimensions of $d=0$, $d=10$, and $d=20$. Each group of positions forms a sinusoid with a different wavelength. In this paper, we use $d=0$ to calculate SPE because the high-frequency sinusoid enables Fast FMVNet V3 to achieve the best performance.

\subsubsection{Details of DAM}
The overview of DAM is depicted in Fig.~\ref{fig:depth_aware_module}. For ease of description, we here assume that the input feature maps are expressed by $\boldsymbol{M} \in \mathbb{R}^{C \times H \times W}$ with channels $C$, height $H$, and width $W$. First, the feature maps pass through a global average (GAP) pooling layer to produce a vector $\boldsymbol{g} \in \mathbb{R}^C$ by the following Eq.~(\ref{eq:global_avg_pool}),
\begin{equation}
	\boldsymbol{g}_k = \text{GAP}(\boldsymbol{M})_k = \frac{1}{H \times W} \sum_{i=0}^{H-1}\sum_{j=0}^{W-1} \boldsymbol{M}_{k,i,j},
	\label{eq:global_avg_pool}
\end{equation}
where $k$, $i$, and $j$ are subscripts. GAP is used to gather the global contextual information in each feature map.

Next, the $\boldsymbol{g \in \mathbb{R}^C}$ and the vector $\boldsymbol{z} \in \mathbb{R}^C$ from above SPE go through a shared multi-layer perceptron (MLP) which is consisted of two fully connected layers and an activation function. Then, both outputs are summed and further scaled to a vector $\boldsymbol{s} \in \mathbb{R}^C$ by a sigmoid function. The process can be presented by the following Eq.~(\ref{eq:scale}),
\begin{equation}
	\boldsymbol{s} = \text{Sigmoid}\left\{\text{MLP}(\boldsymbol{g}) + \text{MLP}(\boldsymbol{z})\right\}.
	\label{eq:scale}
\end{equation} 
By Eq.~(\ref{eq:scale}), the position information is fused with the global contextual information.

Finally, the feature maps $\boldsymbol{M}$ are multiplied by the scale $\boldsymbol{s}$ to output depth-aware feature maps $\boldsymbol{M}^{\prime} \in \mathbb{R}^{C \times H \times W}$. This can be expressed by the following Eq.~(\ref{eq:depth_aware_feature_maps}),
\begin{equation}
	\boldsymbol{M}^{\prime}_{k,:,:} = \boldsymbol{s}_k \times \boldsymbol{M}_{k,:,:}.
	\label{eq:depth_aware_feature_maps}
\end{equation} 
Eq.~(\ref{eq:depth_aware_feature_maps}) indicates that the feature maps $\boldsymbol{M}$ are dynamically recalibrated by considering the global contextual information and position information in each feature map. Therefore, the depth information in the range image is considered.

\begin{figure*}[t]
	\centering
	\includegraphics[width=1.72\columnwidth]{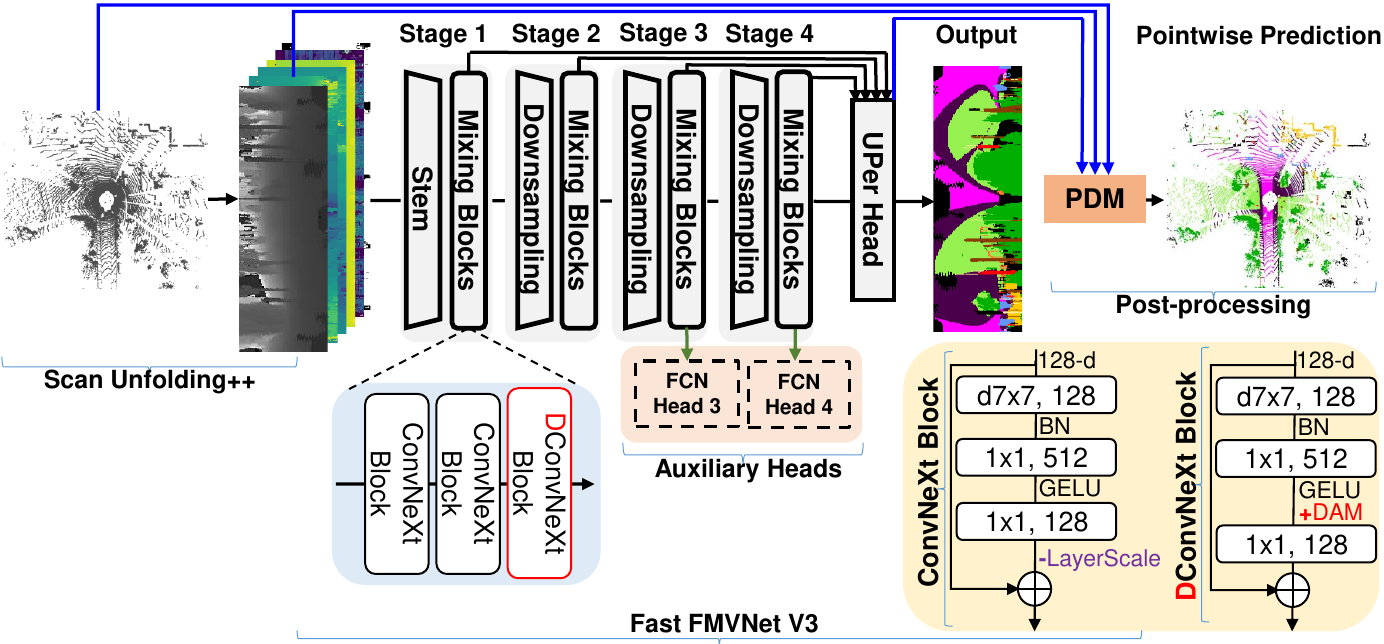}
	\caption{(a) Overview of the introduced Fast FMVNet V3. A point cloud is first projected onto the range image by scan unfolding++~\cite{filling_missing2024}. Then, the range image goes through the backbone of Fast FMVNet V3, which contains four Stages, to output feature maps. Finally, the feature maps pass through the decoder part UPer Head~\cite{upernet2018} and the post-processing module PDM~\cite{pdm2024} to output pointwise predictions. In Fast FMVNet V3, the last ConvNeXt Block in each Stage is the depth-aware ConvNeXt Block (\textcolor{red}{D}ConvNeXt Block) incorporating the proposed depth-aware module (\textcolor{red}{DAM}).}
	\label{fig:fast_fmvnet_v3}  
\end{figure*}

\subsection{Design of Fast FMVNet V3}
The overview of the introduced Fast FMVNet V3 is provided in Fig.~\ref{fig:fast_fmvnet_v3}. Similar to Fast FMVNet V2~\cite{pdm2024}, Fast FMVNet V3 contains four components: pre-processing, encoder, decoder, and post-processing modules. In the pre-processing part, we adopt scan unfolding++ (SU++)~\cite{filling_missing2024} to prepare the range image because compared with the commonly used spherical projection~\cite{rangenet++}, SU++ can keep the shapes and patterns of objects in the range image. The encoder part contains four Stages built by Stem, Downsampling, and Mixing Blocks. The decoder part is the usually used UPer Head~\cite{upernet2018}. In the post-processing part, a trainable pointwise decoder module (PDM)~\cite{pdm2024} is adopted to refine and make the final predictions.

Different from Fast FMVNet V2~\cite{pdm2024}, Fast FMVNet V3 uses the Mixing Blocks in which the last block is the depth-aware ConvNeXt Block (\textcolor{red}{D}ConvNeXt Block) and the rest blocks are ConvNeXt Blocks. Similar to the work~\cite{convnextv223}, \textcolor{red}{D}ConvNeXt Block removes LayerScale~\cite{layerscale21} and adds DAM after GELU. The experimental results will show the superiority of the design of Fast FMVNet V3 regarding the speed and accuracy.

\section{EXPERIMENTS}
In this section, we first describe the experimental settings. Then, ablation study results are reported. Finally, more comparison results on SemanticKITTI~\cite{semantickitti_2019_behley}, nuScenes~\cite{nuscenes_panoptic}, and SemanticPOSS~\cite{semanticposs_2020} datasets are provided.

\subsection{Experimental Settings}
\subsubsection{Datasets} All experiments are conducted on the SemanticKITTI~\cite{semantickitti_2019_behley}, nuScenes~\cite{nuscenes_panoptic}, and SemanticPOSS~\cite{semanticposs_2020} datasets. SemanticKITTI has the sequences \{$00\sim07$ \& $09\sim10$\}, \{$08$\}, and \{$11\sim21$\} as the training, validation, and test datasets, respectively. Besides, only 19 classes are considered in the evaluation under the single scan condition. nuScenes has 28,130 training, 6,019 validation, and 6,008 test samples. Moreover, 16 classes are taken into consideration in experiments. SemanticPOSS has the sequences \{$00, 01, 03, 04, 05$\} and \{$02$\} which serve as the training and test datasets. Also, 14 classes are considered in the evaluation. 

\subsubsection{Training and Testing Settings} During training, Fast FMVNet V3 is trained on the SemanticKITTI, nuScenes, and SemanticPOSS for 50, 80, and 50 epochs, respectively. Also, the corresponding batch sizes are set to 8, 8, and 2, respectively. The optimizer AdamW with a learning rate of 0.002 is adopted. Besides, we use the following data augmentation techniques: random horizontal flipping, random scaling, random rotation, PolarMix~\cite{polarmix_2022}, LaserMix~\cite{lasermix_2023}, and VRCrop~\cite{pdm2024}. Furthermore, we fix all random seeds during training and testing for reproduction and fair comparison. No test-time augmentation and ensemble tricks are utilized during testing to boost the performance for fair comparison. The metrics intersection-over-union (IoU) and mean IoU (mIoU) are used in evaluation.

\subsection{Ablation Study}
In this section, we first study the proposed depth-aware module (DAM). Then, we explore the \textcolor{red}{D}ConvNeXt Block in Fast FMVNet V3.

\begin{table}[t]
	\caption{The impacts of global average pooling (GAP) and sinusoidal positional encoding (SPE) in the depth-aware module (DAM). ``\ding{52}": with the component.}
	\label{tab:gap_pe_dam}
	\centering
	\scalebox{1.0}{
		\begin{tabular}{l|c|l}
			GAP	& SPE  & mIoU (\%) \\ \hline 
			&                      & 68.55~\cite{pdm2024}     \\
			\ding{52}               &                      & 68.32 (\textcolor{red}{-0.23})    \\
			& \ding{52}            & 67.48 (\textcolor{red}{-1.07})    \\
			\ding{52}               & \ding{52}            & \textbf{68.84} (\textcolor{blue}{+0.29})   
	\end{tabular}}
\end{table}

\begin{table}[t]
	\caption{Alternatives to the sinusoidal positional encoding (SPE) in the depth-aware module (DAM). ``$d$" indicates that the positions generated under the dimension $d$ (see Eq.~(\ref{eq:positional_encoding})).}
	\label{tab:pe_others_dam}
	\centering
	\scalebox{1.0}{
		\begin{tabular}{l|c|c|c}
			Alternatives to SPE & Latency & FPS & mIoU (\%) \\ \hline 
			$d=0$ (Ours)       &\multirow{6}*{39.3ms}  & \multirow{6}*{25.5}  & \textbf{68.84}   \\
			$d=5$              &         &     & 67.78     \\
			$d=10$             &         &     & 67.42     \\
			$d=15$             &         &     & 68.11     \\
			$d=20$             &         &     & 67.33     \\
			$d=25$             &         &     & 67.71     \\ \hline
			Learned PE~\cite{learnpe17}         &39.3ms   &25.5 & 67.17  \\
			Adaptive Max Pooling~\cite{cbam18}  &55.0ms   &18.2 & 67.93  \\
			
	\end{tabular}}
\end{table}

\subsubsection{Impacts of GAP and SPE in DAM}
In DAM, global average pooling (GAP) aims to summarize the global contextual information in each feature map. Sinusoidal positional encoding (SPE) is to make the model sensitive to the depth information in the range image. The ablation study results of DAM are reported in Tables~\ref{tab:gap_pe_dam} and~\ref{tab:pe_others_dam}. Here, the baseline result 68.55\% in Table~\ref{tab:gap_pe_dam} is copied from Fast FMVNet V2~\cite{pdm2024}.

In Table~\ref{tab:gap_pe_dam}, we see that only with the combination of GAP and SPE, Fast FMVNet V3 achieves better performance than the counterpart Fast FMVNet V2 (68.84\% (+0.29) \textit{vs.} 68.55\%). Besides, by only applying GAP, the model performance is slightly decreased, probably because of the order of the depth values in the range image. Moreover, by only employing SPE, the model performance is significantly decreased to 67.48\% because each position value cannot effectively represent the global contextual information in the feature map. Considering the importance of GAP, we only explore the alternatives to SPE further in the following experiments.

Table~\ref{tab:pe_others_dam} shows that the positions generated under the dimensions 5, 10, 15, 20, and 25 decrease the model performance. The results indicate that only the high-frequency sinusoid working together with GAP enables the model to effectively perceive the ordered depth information in the range image. Besides, the learned position encoding~\cite{learnpe17} cannot increase the performance, suggesting that explicitly modelling the interdependent relationships among the channels for the ordered depth values is required. Moreover, replacing our SPE with the adaptive max pooling~\cite{cbam18} results in inferior performance and high inference latency. The potential explanation is that the maximum value in each feature map is likely to be related to the object distant from the LiDAR sensor, and the value cannot effectively represent other objects in each feature map.

\begin{table}[t]
	\caption{The settings of the \textcolor{red}{D}ConvNeXt Block in the Fast FMVNet V3 architecture. ``CB": ConvNeXt Block.}
	\label{tab:dconvnext_block}
	\centering
	\scalebox{1.0}{
		\begin{tabular}{l|c|c|c}
			Settings                             & Latency & FPS  & mIoU (\%) \\ \hline 
			Replace all CB                       & 41.4ms  & 24.2 & 68.17     \\
			Replace the last 2 CB in each Stage  & 40.1ms  & 24.9 & 68.32     \\
			Replace the last 1 CB in each Stage  & 39.3ms  & 25.5 & \textbf{68.84}     \\
	\end{tabular}}
\end{table}

\subsubsection{Where should the DConvNeXt Block be placed?}
In Fast FMVNet V3, Stages 1, 2, 3, and 4 store [3, 4, 6, 3] blocks, respectively. Hence, we consider three settings: replacing all ConvNeXt Blocks with the introduced \textcolor{red}{D}ConvNeXt Blocks, replacing the last two ConvNeXt Blocks in each Stage with \textcolor{red}{D}ConvNeXt Blocks, and replacing the last one ConvNeXt Block in each Stage with \textcolor{red}{D}ConvNeXt Block.

Table~\ref{tab:dconvnext_block} demonstrates that only the last ConvNeXt Block in each Stage replaced with \textcolor{red}{D}ConvNeXt Block leads to the best performance in terms of the inference speed and mIoU score. 

\begin{table}[t]
	\caption{Comparisons among different attention modules. ``CBAM": Convolutional Block Attention Module~\cite{cbam18}. ``SSSC", ``CCCS", ``SSCC", and ``CCSS": the settings of the spatial (Pooling) and channel (Depth-Aware Module) attention modules in four Fast FMVNet V3 Stages.}
	\label{tab:dam_attention_modules}
	\centering
	\scalebox{0.93}{
		\begin{tabular}{l|c|c|c}
			Attention Modules                                 & Latency & FPS  & mIoU (\%)        \\ \hline 
			Depth-Aware Module (Ours)                         & 39.3ms  & 25.5 & \textbf{68.84}   \\ 
			
			Squeeze-and-Excitation block~\cite{senet18}       & 39.2ms  & 25.5 & 67.25            \\
			Global Response Normalization~\cite{convnextv223} & 41.1ms  & 24.3 & \textcolor{purple}{68.70}            \\
			CBAM-Channel Attention~\cite{cbam18}              & 55.0ms  & 18.2 & 67.93            \\ \hline
			
			CBAM-Spatial Attention~\cite{cbam18}              & 39.0ms  & 25.6 & 67.66            \\ 
			Pooling~\cite{metaformer22}                       & 41.8ms  & 23.9 & 68.30            \\ 
			Multi-Scale Convolutional Attention~\cite{segnext22}&45.4ms & 22.0 & 67.22            \\
			Focal Modulation~\cite{focalnet22}                & 42.9ms  & 23.3 & 67.81            \\ \hline
			
			Depth-Aware Module \& Pooling (SSSC)              & 41.6ms  & 24.0 & \textcolor{purple}{68.72}            \\
			Depth-Aware Module \& Pooling (CCCS)              & 40.0ms  & 25.0 & \textcolor{purple}{68.83}            \\
			Depth-Aware Module \& Pooling (SSCC)              & 41.5ms  & 24.1 & 67.58            \\
			Depth-Aware Module \& Pooling (CCSS)              & 39.4ms  & 25.4 & 67.04            \\
	\end{tabular}}
\end{table}

\begin{table*}[t]
	\caption{Comparisons among range image-based models on the SemanticKITTI test set in terms of IoU and mIoU scores (\%). ``$\dagger$" means that \textbf{test-time augmentation (TTA)} is applied to boost the performance. Note that \textbf{none of tricks} are employed to increase our results further. \textbf{Bold} font: the best results. \underline{Underline}: the second-best results.}
	\label{tab:64x2048_kitti_test_results}
	\centering
	\scalebox{0.78}{
		\begin{tabular}{l|c|l|c|c|c|c|c|c|c|c|c|c|c|c|c|c|c|c|c|c|c}
			Models	& Years & mIoU &\rotatebox{90}{Car} &\rotatebox{90}{Bicycle} &\rotatebox{90}{Motorcycle} &\rotatebox{90}{Truck} &\rotatebox{90}{Other-vehicle} &\rotatebox{90}{Person} &\rotatebox{90}{Bicyclist} &\rotatebox{90}{Motorcyclist} &\rotatebox{90}{Road} &\rotatebox{90}{Parking} &\rotatebox{90}{Sidewalk} &\rotatebox{90}{Other-ground} &\rotatebox{90}{Building} &\rotatebox{90}{Fence} &\rotatebox{90}{Vegetation} &\rotatebox{90}{Trunk} &\rotatebox{90}{Terrain} &\rotatebox{90}{Pole} &\rotatebox{90}{Traffic-sign}  \\ \hline 
			
			SqueezeSeg~\cite{squeezeseg} & 2018 &30.8 &68.3 &18.1 &5.1 &4.1 &4.8 &16.5 &17.3 &1.2 &84.9 &28.4 &54.7 &4.6 &61.5 &29.2 &59.6 &25.5 &54.7 &11.2 &36.3  \\
			SqueezeSegV2~\cite{squeezesegv2} & 2019 &39.7 &81.8 &18.5 &17.9 &13.4 &14.0 &20.1 &25.1 &3.9 &88.6 &45.8 &67.6 &17.7 &73.7 &41.1 &71.8 &35.8 &60.2 &20.2 &36.3  \\ 
			
			RangeNet21~\cite{rangenet++}& 2019 &47.4 &85.4 &26.2 &26.5 &18.6 &15.6 &31.8 &33.6 &4.0 &91.4 &57.0 &74.0 &26.4 &81.9 &52.3 &77.6 &48.4 &63.6 &36.0 &50.0 \\ 
			
			RangeNet53++~\cite{rangenet++}& 2019 &52.2 &91.4 &25.7 &34.4 &25.7 &23.0 &38.3 &38.8 &4.8 &91.8 &65.0 &75.2 &27.8 &87.4 &58.6 &80.5 &55.1 &64.6 &47.9 &55.9 \\
			
			SqSegV3-21~\cite{squeezesegv3_2020}& 2020 &51.6 &89.4 &33.7 &34.9 &11.3 &21.5 &42.6 &44.9 &21.2 &90.8 &54.1 &73.3 &23.2 &84.8 &53.6 &80.2 &53.3 &64.5 &46.4 &57.6 \\ 
			SqSegV3-53~\cite{squeezesegv3_2020}& 2020 &55.9 &92.5 &38.7 &36.5 &29.6 &33.0 &45.6 &46.2 &20.1 &91.7 &63.4 &74.8 &26.4 &89.0 &59.4 &82.0 &58.7 &65.4 &49.6 &58.9 \\ 
			
			FIDNet~\cite{fidnet_2021}& 2021 &59.5 &93.9 &54.7 &48.9 &27.6 &23.9 &62.3 &59.8 &23.7 &90.6 &59.1 &75.8 &26.7 &88.9 &60.5 &84.5 &64.4 &69.0 &53.3 &62.8  \\ 
			
			CENet$\dagger$~\cite{cenet_2022}& 2022 &64.7 &91.9 &58.6 &50.3 &40.6 &42.3 &68.9 &65.9 &43.5 &90.3 &60.9 &75.1 &31.5 &91.0 &66.2 &84.5 &69.7 &70.0 &61.5 &67.6 \\ 
			
			RangeViT~\cite{rangevit_2023} & 2023 & 64.0 & 95.4 &55.8 &43.5 &29.8 &42.1 &63.9 &58.2 &38.1 &\textbf{93.1} &70.2 &\underline{80.0} &32.5 &92.0 &69.0 &\underline{85.3} &70.6 &\underline{71.2} &60.8 &64.7 \\ 
			
			RangeFormer~\cite{rangeformer_2023}& 2023 &\textcolor{blue}{69.5} &\textcolor{blue}{94.7} &\textcolor{blue}{60.0} &\underline{\textcolor{blue}{69.7}} &\underline{\textcolor{blue}{57.9}} &\underline{\textcolor{blue}{64.1}} &\textcolor{blue}{72.3} &\underline{\textcolor{blue}{72.5}} &\underline{\textcolor{blue}{54.9}} &\textcolor{blue}{90.3} &\textcolor{blue}{69.9} &\textcolor{blue}{74.9} &\underline{\textcolor{blue}{38.9}} &\textcolor{blue}{90.2} &\textcolor{blue}{66.1} &\textcolor{blue}{84.1} &\textcolor{blue}{68.1} &\textcolor{blue}{70.0} &\textcolor{blue}{58.9} &\textcolor{blue}{63.1} \\

			RangeFormer$\dagger$~\cite{rangeformer_2023}& 2023 &\textbf{73.3} &\underline{96.7} &\textbf{69.4} &\textbf{73.7} &\textbf{59.9} &\textbf{66.2} &\textbf{78.1} &\textbf{75.9} &\textbf{58.1} &92.4 &73.0 &78.8 &\textbf{42.4} &92.3 &\textbf{70.1} &\textbf{86.6} &\textbf{73.3} &\textbf{72.8} &\textbf{66.4} &66.6 \\

			FMVNet~\cite{filling_missing2024} &2024 &68.0 &96.6 & 63.4 & 60.9 & 42.1 & 55.5 &\underline{75.6} & 70.7 & 26.1 & 92.5 &\textbf{73.8} &79.3 & 37.7 &92.3 &69.3 & 85.2 &\underline{71.4} & 69.7 &\underline{63.0} &66.8 \\ 
			
			Fast FMVNet V2~\cite{pdm2024} &2024 &68.1 &96.1 &\underline{64.2} &66.4 &42.3 &49.1 &74.9 &67.9 &43.5 &92.6 &72.0 &79.1 &34.0 &\underline{92.5} &68.5 &84.5 & 68.6 &69.5 &59.1 &\underline{68.5} \\ 	
			
			Fast FMVNet V3  &2025 &\underline{\textcolor{blue}{69.6}} &\textbf{\textcolor{blue}{96.9}} & \textcolor{blue}{61.5} & \textcolor{blue}{64.9} & \textcolor{blue}{51.4} & \textcolor{blue}{57.5} & \textcolor{blue}{74.9} & \textcolor{blue}{70.2} & \textcolor{blue}{42.8} & \underline{\textcolor{blue}{93.0}} & \underline{\textcolor{blue}{73.1}} & \textbf{\textcolor{blue}{80.3}} & \textcolor{blue}{37.3} & \textbf{\textcolor{blue}{92.5}} & \underline{\textcolor{blue}{69.8}} & \textcolor{blue}{84.5} & \textcolor{blue}{71.2} & \textcolor{blue}{68.7} & \textcolor{blue}{61.0} & \textbf{\textcolor{blue}{70.8}}\\
	\end{tabular}}
\end{table*}

\begin{table*}[h]
	\caption{Comparison results among range image-based models on the nuScenes validation dataset regarding IoU and mIoU scores (\%). Note that \textbf{none of test-time augmentation and ensemble tricks} are applied to our results. ``Constr. Veh.": ``Construction Vehicle"; ``Drive. Sur.": ``Driveable Surface"; \textbf{Bold} font: the best results. \underline{Underline}: the second-best results.}
	\label{tab:32x1088_nuscenes_val_results}
	\centering
	\scalebox{0.91}{
		\begin{tabular}{l|l|c|c|c|c|c|c|c|c|c|c|c|c|c|c|c|c}
			Models	 &mIoU &\rotatebox{90}{Barrier} &\rotatebox{90}{Bicycle} &\rotatebox{90}{Bus} &\rotatebox{90}{Car} &\rotatebox{90}{Constr. Veh.} &\rotatebox{90}{Motorcycle} &\rotatebox{90}{Pedestrian} &\rotatebox{90}{Traffic Cone} &\rotatebox{90}{Trailer} &\rotatebox{90}{Truck} &\rotatebox{90}{Drive. Sur.} &\rotatebox{90}{Other Flat} &\rotatebox{90}{Sidewalk} &\rotatebox{90}{Terrain} &\rotatebox{90}{Manmade} &\rotatebox{90}{Vegetation}  \\ \hline
			RangeNet53++~\cite{rangenet++,filling_missing2024} &71.1 &58.5 &38.1 &90.0  &84.0 &46.1 &80.1 &62.3 &42.3 &62.4 &80.9 &96.5 &73.7 &75.1 &74.2 &87.6 &86.0 \\ 
			FIDNet~\cite{fidnet_2021,filling_missing2024} &73.5 &59.5 &44.2 &88.4 &84.6 &48.1 &84.0 &70.4 &59.9 &65.7 &78.0 &96.5 &71.6 &74.7 &75.1 &88.7 &87.3 \\ 
			CENet~\cite{cenet_2022,filling_missing2024}  &73.4 &60.2 &43.0 &88.0 &85.0 &53.6 &70.4 &71.0 &62.5 &65.6 &80.1 &96.6 &72.3 &74.9 &75.1 &89.1 &87.7 \\

			RangeViT~\cite{rangevit_2023} &75.2 &75.5 &40.7 &88.3 &90.1 &49.3 &79.3 &77.2 &66.3 &65.2 &80.0 &96.4 &71.4 &73.8 &73.8 &89.9 &87.2 \\ 
			
			RangeFormer+STR~\cite{rangeformer_2023}&\underline{77.1} &\underline{76.0} &44.7 &94.2 &\underline{92.2} &54.2 &82.1 &76.7 &\textbf{69.3} &61.8 &83.4 &96.7 &\textbf{75.7} &75.2 &\underline{75.4} &88.8 &87.3 \\ 
			RangeFormer~\cite{rangeformer_2023}&\textbf{78.1} &\textbf{78.0} &45.2 &94.0 &\textbf{92.9} &\underline{58.7} &83.9 &\textbf{77.9} &69.1 &63.7 &\underline{85.6} &96.7 &74.5 &75.1 &75.3 &89.1 &87.5 \\ 
			
			FMVNet~\cite{filling_missing2024}&76.8 &61.1 &\textbf{49.5} &94.7 &86.8 &\textbf{59.6} &71.1 &\underline{77.2} &\underline{69.1} &\textbf{70.9} &\textbf{85.6} &\underline{96.9} &\underline{75.0} &\textbf{76.5} &\textbf{75.8} &90.1 &88.4 \\ 
			
			Fast FMVNet~\cite{filling_missing2024}&76.0 &60.3 &\underline{45.8} &\underline{95.1} &86.7 &54.7 &\textbf{85.7} &74.0 &66.2 &67.1 &83.5 &96.7 &72.7 &75.1 &74.8 &89.8 &88.3 \\ 
			
			Fast FMVNet V2~\cite{pdm2024}&76.1 &64.8 &41.0 &\textbf{95.3} &88.6 &54.5 &84.9 &72.5 &62.2 &\underline{67.1} &85.5 &\textbf{96.9} &74.8 &\underline{75.5} &75.1 &\textbf{90.4} &\textbf{88.8} \\ 
			
			Fast FMVNet V3 &76.6 &75.9 &41.1 &94.8 &92.1 &54.4 &\underline{85.2} &73.3 &62.9 &64.5 &84.5 &96.7 &72.2 &75.0 &74.2 &\underline{90.1} &\underline{88.7}
	\end{tabular}}
\end{table*}

\subsection{Comparison with Attention Modules}
The proposed depth-aware module (DAM) can be seen as an attention component. Here, we compare DAM with several attention modules. The experimental results are provided in Table~\ref{tab:dam_attention_modules}.

Table~\ref{tab:dam_attention_modules} shows that among all channel attention modules, \textit{i.e.}, depth-aware module, squeeze-and-excitation block~\cite{senet18}, global response normalization~\cite{convnextv223}, and CBAM-channel attention~\cite{cbam18}, the proposed DAM enables Fast FMVNet V3 to achieve the best performance. Besides, DAM surpasses all spatial attention modules, such as CBAM-spatial attention~\cite{cbam18}, pooling~\cite{metaformer22}, multi-scale convolutional attention~\cite{segnext22}, and focal modulation~\cite{focalnet22}. Moreover, combining DAM and pooling with the settings of ``SSSC" and ``CCCS" enables Fast FMVNet V3 to obtain competitive performance. Here, ``SSSC" means that the last block in the first three Stages is replaced with the block incorporating the spatial attention module, namely pooling, and the last block in the last Stage is replaced with the block containing the channel attention component, namely our DAM. ``CCCS" is similar to ``SSSC". The comparison results validate the effectiveness of the proposed DAM.

\subsection{Comparison Results on SemanticKITTI}
The comparison results among range image-based models on the SemanticKITTI~\cite{semantickitti_2019_behley} test dataset are reported in Table~\ref{tab:64x2048_kitti_test_results}. We see that Fast FMVNet V3 significantly surpasses Fast FMVNet V2, \textit{i.e.}, 69.6\% (+1.5) \textit{vs.} 68.1\%. The performance gain can be attributed to the proposed depth-aware module (DAM) and architecture design. Besides, without test-time augmentation (TTA) to boost performance, Fast FMVNet V3 outperforms RangeFormer~\cite{rangeformer_2023} regarding the mIoU score (\textit{i.e.}, 69.6\% \textit{vs.} 69.5\%). In addition, although Fast FMVNet V3 is inferior to the boosted RangeFormer$\dagger$, applying TTA for RangeFormer$\dagger$ leads to unacceptable inference latency~\cite{empir24}. By contrast, Fast FMVNet V3 can run at high speed, reaching 25.5 frames per second (see Table~\ref{tab:time_models}). The above experimental results validate the effectiveness of the proposed DAM and Fast FMVNet V3.

\subsection{Comparison Results on nuScenes}
The comparison results between Fast FMVNet V3 and the existing range image-based models on the nuScenes~\cite{nuscenes_panoptic} validation dataset are provided in Table~\ref{tab:32x1088_nuscenes_val_results}. We see that Fast FMVNet V3 obtains better performance than its counterpart Fast FMVNet V2 (76.6\% (+0.5) \textit{vs.} 76.1\%). The improved performance can be attributed to the proposed depth-aware module (DAM) and the architecture design. Besides, Fast FMVNet V3 is slightly inferior to FMVNet~\cite{filling_missing2024} regarding the mIoU score, but Fast FMVNet V3 is significantly faster than FMVNet and has only 4.5M model parameters (see Table~\ref{tab:time_models}). In addition, Fast FMVNet V3 is inferior to RangeFormer+STR and RangeFormer. This is because following the experimental settings in the works~\cite{filling_missing2024,pdm2024}, we remove all invalid points during training but consider the invalid points during testing, and there are at least 5.7\% erroneously annotated points in the nuScenes validation dataset. 

\begin{table*}[t]
	\caption{Comparison results among range image-based models on the SemanticPOSS test dataset (sequence $\left\{02\right\}$) in terms of IoU and mIoU scores (\%). Note that \textbf{none of test-time augmentation and ensemble tricks} are employed to boost our results. \textbf{Bold} font: the best results. \underline{Underline}: the second-best results.}
	\label{tab:64x2048_poss_test_results}
	\centering
	\scalebox{0.86}{
		\begin{tabular}{l|l|c|c|c|c|c|c|c|c|c|c|c|c|c}
			Models &mIoU &People &Rider &Car &Trunk &Plants &Traffic Sign &Pole &Trashcan &Building &Cone/Stone &Fence &Bike &Ground \\ \hline 
			SqueezeSeg~\cite{squeezeseg}     &18.9 &14.2 &1.0  &13.2 &10.4 &28.0 &5.1  &5.7  &2.3  &43.6 &0.2  &15.6 &31.0 &75.0 \\  
			SqueezeSegV2~\cite{squeezesegv2} &30.0 &48.0 &9.4  &48.5 &11.3 &50.1 &6.7  &6.2  &14.8 &60.4 &5.2  &22.1 &36.1 &71.3 \\ 
			MINet~\cite{minet_2021}          &43.2 &62.4 &12.1 &63.8 &22.3 &68.6 &16.7 &30.1 &28.9 &75.1 &28.6 &32.2 &44.9 &76.3 \\ 
			
			RangeNet53++~\cite{rangenet++,filling_missing2024} &51.4 &74.6 &22.6 &79.8 &\textbf{26.9} &71.3 &21.3 &28.2 &31.6 &77.5 &49.3 &51.7 &54.9 &77.9 \\  
			FIDNet~\cite{fidnet_2021,filling_missing2024}      &53.5 &78.5 &29.6 &79.0 &25.8 &71.4 &23.3 &32.8 &38.4 &79.2 &49.4 &54.4 &55.9 &78.2 \\ 
			CENet~\cite{cenet_2022,filling_missing2024}        &54.3 &78.1 &29.0 &83.0 &26.4 &70.5 &22.9 &\underline{33.6} &36.6 &79.2 &\textbf{58.1} &53.1 &56.2 &79.6 \\ 
			
			FMVNet~\cite{filling_missing2024} &55.1 &80.0 &29.9 &\underline{84.2} &26.2 &73.4 &\underline{25.5} &31.4 &34.9 &82.5 &\underline{55.0} &\underline{55.9} &56.4 &\underline{80.9} \\ 
			
			Fast FMVNet~\cite{filling_missing2024} &54.7 &80.1 &29.2 &83.9 &\underline{26.7} &73.1 &24.8 &32.7 &\underline{40.8} &81.4 &48.8 &54.8 &56.3 &78.4 \\ 
			
			Fast FMVNet V2~\cite{pdm2024}&\underline{55.3} &\textbf{80.2} &\underline{30.0} &\textbf{85.0} &25.8 &\underline{73.6} &23.7 &31.6 &37.9 &\underline{82.8} &54.9 &55.7 &\underline{57.1} &\textbf{81.1} \\ 
			
			Fast FMVNet V3 &\textbf{56.1} &\underline{80.1} &\textbf{31.9} &83.9 &26.1 &\textbf{74.2} &\textbf{25.8} &\textbf{34.2} &\textbf{41.2} &\textbf{83.4}&53.2 &\textbf{56.1} &\textbf{58.4} &80.6
	\end{tabular}}
\end{table*}

\subsection{Comparison Results on SemanticPOSS}
The comparison results between Fast FMVNet V3 and other range image-based models on the SemanticPOSS~\cite{semanticposs_2020} test dataset are provided in Table~\ref{tab:64x2048_poss_test_results}. We see that Fast FMVNet V3 largely outperforms the counterpart Fast FMVNet V2 (\textit{i.e.}, 56.1\% (+0.8) \textit{vs.} 55.3\%) and other models. The performance gain can validate the effectiveness of the proposed depth-aware module (DAM) and the superiority of the architecture design. 

\begin{table}[t]
	\caption{Time comparisons among various models regarding model parameters (Params.), inference latency, frames per second (FPS), and mIoU scores (\%) on the SemanticKITTI~\cite{semantickitti_2019_behley} validation dataset. ``$^\ast$": results from the work~\cite{pdm2024}.}
	\label{tab:time_models}
	\centering
	\scalebox{0.93}{
		\begin{tabular}{l|c|r|r|l|c}
			Methods                                & Years &Params. & Latency &FPS   & mIoU  \\ \hline 
			MinkUNet~\cite{minkowski2019}          & 2019  &21.7M   & 48.4ms  &20.7  & 61.1  \\  
			Cylinder3D~\cite{cylindrical3d2021}    & 2021  &56.3M   & 71.5ms  &13.3  & 65.9  \\ 
			UniSeg 0.2$\times$~\cite{openpcseg2023}& 2023  &28.8M   & 84.6ms  &11.8  & 67.0  \\ 
			UniSeg 1.0$\times$~\cite{openpcseg2023}& 2023  &147.6M  & 145.0ms &6.9   &\textbf{71.3}  \\ 
			
			RangeNet53$^\ast$~\cite{rangenet++,pdm2024} & 2019  &50.4M   & 30.5ms  &32.8  & 64.8  \\ 
			FIDNet$^\ast$~\cite{fidnet_2021,pdm2024}    & 2021  &6.1M    & 32.3ms  &31.0  & 66.2  \\ 
			CENet$^\ast$~\cite{cenet_2022,pdm2024}      & 2022  &6.8M    & 31.8ms  &31.5  & 66.3  \\ 
			RangeFormer~\cite{rangeformer_2023,filling_missing2024} & 2023  &24.3M   & 90.3ms  &11.1  & 67.6  \\
			
			FMVNet~\cite{filling_missing2024}      & 2024    & 59.3M   & 96.1ms  & 10.4 &\underline{69.0} \\ 
			
			Fast FMVNet~\cite{filling_missing2024} & 2024    &4.3M     & 20.8ms  &48.1  & 67.9  \\ 
			Fast FMVNet V2~\cite{pdm2024}          & 2024    &4.4M     & 37.7ms  &26.5  & 68.6 \\ 
			Fast FMVNet V3 (Ours)                        & 2025    &4.5M     & 39.3ms  &25.5  & 68.8 \\
	\end{tabular}}
\end{table}

\begin{figure}[t]
	\centering
	\includegraphics[width=1.0\columnwidth]{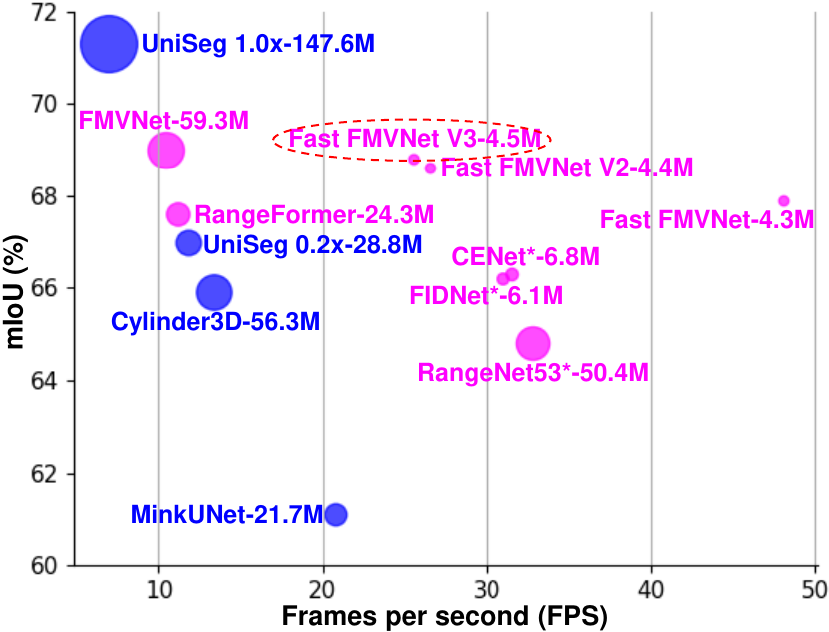}
	\caption{Comparisons among state-of-the-art models regarding model parameters, frames per second (FPS), and mIoU scores (\%) on the SemanticKITTI~\cite{semantickitti_2019_behley} validation dataset. \textcolor[rgb]{0,0,1}{Blue color}: voxel-based models. \textcolor[rgb]{1,0,1}{Pink color}: range image-based models. Circle size: the number of learned model parameters.}
	\label{fig:speed_miou_kitti}  
\end{figure}

\subsection{Time Comparison}
Fast FMVNet V3 aims to become a fast, lightweight, and high accuracy point cloud segmentation (PCS) model. The comprehensive comparisons between Fast FMVNet V3 and exiting voxel- and range image-based models are provided in Table~\ref{tab:time_models} and Fig.~\ref{fig:speed_miou_kitti}. We see that Fast FMVNet V3 only has 4.5M model parameters but can obtain the 68.8\% mIoU score with 25.5 frames per second (FPS). Fig.~\ref{fig:speed_miou_kitti} shows that Fast FMVNet V3 achieves state-of-the-art performance in terms of the speed-accuracy trade-off. Besides, Fast FMVNet V3 is slower than Fast FMVNet V2~\cite{pdm2024} by only one FPS, indicating the negligible computational cost of the proposed depth-aware module (DAM). In addition, compared with FMVNet~\cite{filling_missing2024} and UniSeg$1.0\times$~\cite{openpcseg2023}, Fast FMVNet V3 is inferior to them regarding the mIoU score but can run 2.5$\times$ and 3.7$\times$ faster than them. The comparison results demonstrate the superiority of the introduced DAM and Fast FMVNet V3 architecture.

\section{CONCLUSION}
In this paper, we find that existing range image-based models cannot effectively learn the implicit but ordered depth information inherent in the range image and achieve inferior performance. To fully utilize the ordered depth values, we propose a depth-aware module (DAM) and introduce Fast FMVNet V3. DAM contains a global contextual information extraction part and positional encoding part, effectively perceiving the ordered depth information in the range image. Fast FMVNet V3 integrates DAM into the last block in each stage, achieving high performance with negligible computational cost. The experimental results on the SemanticKITTI, nuScenes, and SemanticPOSS demonstrate the effectiveness of the introduced DAM and Fast FMVNet V3. In the future, we plan to apply Fast FMVNet V3 to range image-based semantic SLAM systems.

%A conclusion section is not required. Although a conclusion may review the main points of the paper, do not replicate the abstract as the conclusion. A conclusion might elaborate on the importance of the work or suggest applications and extensions. 


%%%%%%%%%%%%%%%%%%%%%%%%%%%%%%%%%%%%%%%%%%%%%%%%%%%%%%%%%%%%%%%%%%%%%%%%%%%%%%%%

%%%%%%%%%%%%%%%%%%%%%%%%%%%%%%%%%%%%%%%%%%%%%%%%%%%%%%%%%%%%%%%%%%%%%%%%%%%%%%%%

\bibliographystyle{IEEEtran}
\bibliography{root.bib}

\begin{thebibliography}{10}
\providecommand{\url}[1]{#1}
\csname url@rmstyle\endcsname
\providecommand{\newblock}{\relax}
\providecommand{\bibinfo}[2]{#2}
\providecommand\BIBentrySTDinterwordspacing{\spaceskip=0pt\relax}
\providecommand\BIBentryALTinterwordstretchfactor{4}
\providecommand\BIBentryALTinterwordspacing{\spaceskip=\fontdimen2\font plus
\BIBentryALTinterwordstretchfactor\fontdimen3\font minus
  \fontdimen4\font\relax}
\providecommand\BIBforeignlanguage[2]{{%
\expandafter\ifx\csname l@#1\endcsname\relax
\typeout{** WARNING: IEEEtran.bst: No hyphenation pattern has been}%
\typeout{** loaded for the language `#1'. Using the pattern for}%
\typeout{** the default language instead.}%
\else
\language=\csname l@#1\endcsname
\fi
#2}}

\bibitem{semantickitti_2019_behley}
J.~Behley, M.~Garbade, A.~Milioto, J.~Quenzel, S.~Behnke, C.~Stachniss, and
  J.~Gall, ``Semantickitti: A dataset for semantic scene understanding of lidar
  sequences,'' in \emph{ICCV}, 2019, pp. 9297--9307.

\bibitem{nuscenes_panoptic}
W.~K. Fong, R.~Mohan, J.~V. Hurtado, L.~Zhou, H.~Caesar, O.~Beijbom, and
  A.~Valada, ``Panoptic nuscenes: A large-scale benchmark for lidar panoptic
  segmentation and tracking,'' \emph{IEEE RAL}, vol.~7, no.~2, pp. 3795--3802,
  2022.

\bibitem{semanticposs_2020}
Y.~Pan, B.~Gao, J.~Mei, S.~Geng, C.~Li, and H.~Zhao, ``Semanticposs: A point
  cloud dataset with large quantity of dynamic instances,'' in \emph{IEEE IV},
  2020, pp. 687--693.

\bibitem{suma++_2019}
X.~Chen, A.~Milioto, E.~Palazzolo, P.~Giguère, J.~Behley, and C.~Stachniss,
  ``Suma++: Efficient lidar-based semantic slam,'' in \emph{IROS}, 2019, pp.
  4530--4537.

\bibitem{sa_loam_2021}
L.~Li, X.~Kong, X.~Zhao, W.~Li, F.~Wen, H.~Zhang, and Y.~Liu, ``Sa-loam:
  Semantic-aided lidar slam with loop closure,'' in \emph{ICRA}, 2021, pp.
  7627--7634.

\bibitem{fidnet_2021}
Y.~Zhao, L.~Bai, and X.~Huang, ``Fidnet: Lidar point cloud semantic
  segmentation with fully interpolation decoding,'' in \emph{IROS}, 2021, pp.
  4453--4458.

\bibitem{pdm2024}
B.~Chen, C.~Gong, A.~Tikanmäki, and J.~Röning, ``Trainable pointwise decoder
  module for point cloud segmentation,'' in \emph{arXiv}, 2024.

\bibitem{minkowski2019}
C.~Choy, J.~Gwak, and S.~Savarese, ``4d spatio-temporal convnets: Minkowski
  convolutional neural networks,'' in \emph{CVPR}, 2019, pp. 3070--3079.

\bibitem{spvnas_2020}
H.~Tang, Z.~Liu, S.~Zhao, Y.~Lin, J.~Lin, H.~Wang, and S.~Han, ``Searching
  efficient 3d architectures with sparse point-voxel convolution,'' in
  \emph{ECCV}, 2020, pp. 685--702.

\bibitem{waffleiron23}
G.~Puy, A.~Boulch, and R.~Marlet, ``Using a waffle iron for automotive point
  cloud semantic segmentation,'' in \emph{ICCV}, 2023, pp. 3356--3366.

\bibitem{pointtransv32024}
X.~Wu, L.~Jiang, P.-S. Wang, Z.~Liu, X.~Liu, Y.~Qiao, W.~Ouyang, T.~He, and
  H.~Zhao, ``Point transformer v3: Simpler, faster, stronger,'' in \emph{CVPR},
  2024, pp. 4840--4851.

\bibitem{resnet_2016}
K.~He, X.~Zhang, S.~Ren, and J.~Sun, ``Deep residual learning for image
  recognition,'' in \emph{CVPR}, 2016, pp. 770--778.

\bibitem{convnext2022}
Z.~Liu, H.~Mao, C.-Y. Wu, C.~Feichtenhofer, T.~Darrell, and S.~Xie, ``A convnet
  for the 2020s,'' in \emph{CVPR}, 2022, pp. 11\,966--11\,976.

\bibitem{senet18}
J.~Hu, L.~Shen, and G.~Sun, ``Squeeze-and-excitation networks,'' in
  \emph{CVPR}, 2018, pp. 7132--7141.

\bibitem{convnextv223}
S.~Woo, S.~Debnath, R.~Hu, X.~Chen, Z.~Liu, I.~S. Kweon, and S.~Xie, ``Convnext
  v2: Co-designing and scaling convnets with masked autoencoders,'' in
  \emph{CVPR}, 2023, pp. 16\,133--16\,142.

\bibitem{kitti12}
A.~Geiger, P.~Lenz, and R.~Urtasun, ``Are we ready for autonomous driving? the
  kitti vision benchmark suite,'' in \emph{CVPR}, 2012, pp. 3354--3361.

\bibitem{squeezeseg}
B.~Wu, A.~Wan, X.~Yue, and K.~Keutzer, ``Squeezeseg: Convolutional neural nets
  with recurrent crf for real-time road-object segmentation from 3d lidar point
  cloud,'' in \emph{ICRA}, 2018, pp. 1887--1893.

\bibitem{squeezenet_2016}
F.~N. Iandola, S.~Han, M.~W. Moskewicz, K.~Ashraf, W.~J. Dally, and K.~Keutzer,
  ``Squeezenet: Alexnet-level accuracy with 50x fewer parameters and
  \ensuremath{<}0.5{MB} model size,'' in \emph{arXiv}, 2016.

\bibitem{rangenet++}
A.~Milioto, I.~Vizzo, J.~Behley, and C.~Stachniss, ``Rangenet ++: Fast and
  accurate lidar semantic segmentation,'' in \emph{IROS}, 2019, pp. 4213--4220.

\bibitem{yolov3_2018}
J.~Redmon and A.~Farhadi, ``Yolov3: An incremental improvement,'' in
  \emph{arXiv}, 2018.

\bibitem{cenet_2022}
H.~Cheng, X.~Han, and G.~Xiao, ``Cenet: Toward concise and efficient lidar
  semantic segmentation for autonomous driving,'' in \emph{ICME}, 2022, pp.
  01--06.

\bibitem{rangevit_2023}
A.~Ando, S.~Gidaris, A.~Bursuc, G.~Puy, A.~Boulch, and R.~Marlet, ``Rangevit:
  Towards vision transformers for 3d semantic segmentation in autonomous
  driving,'' in \emph{CVPR}, 2023, pp. 5240--5250.

\bibitem{vit_iclr_2021}
A.~Dosovitskiy, L.~Beyer, A.~Kolesnikov, D.~Weissenborn, X.~Zhai,
  T.~Unterthiner, M.~Dehghani, M.~Minderer, G.~Heigold, S.~Gelly, J.~Uszkoreit,
  and N.~Houlsby, ``An image is worth 16x16 words: Transformers for image
  recognition at scale,'' in \emph{ICLR}, 2021.

\bibitem{filling_missing2024}
B.~Chen, C.~Gong, and J.~Röning, ``Filling missing values matters for range
  image-based point cloud segmentation,'' \emph{IEEE TIV}, pp. 1--15, 2024.

\bibitem{segnext22}
M.-H. Guo, C.-Z. Lu, Q.~Hou, Z.-N. Liu, M.-M. Cheng, and S.-M. Hu, ``Segnext:
  rethinking convolutional attention design for semantic segmentation,'' in
  \emph{NeurIPS}, 2022, pp. 1140 -- 1156.

\bibitem{metaformer22}
W.~Yu, M.~Luo, P.~Zhou, C.~Si, Y.~Zhou, X.~Wang, J.~Feng, and S.~Yan,
  ``Metaformer is actually what you need for vision,'' in \emph{CVPR}, 2022,
  pp. 10\,819--10\,829.

\bibitem{focalnet22}
J.~Yang, C.~Li, X.~Dai, and J.~Gao, ``Focal modulation networks,'' in
  \emph{NeurIPS}, 2022, pp. 4203--4217.

\bibitem{rangedet21}
L.~Fan, X.~Xiong, F.~Wang, N.~Wang, and Z.~Zhang, ``Rangedet: In defense of
  range view for lidar-based 3d object detection,'' in \emph{ICCV}, 2021, pp.
  2898--2907.

\bibitem{cbam18}
S.~Woo, J.~Park, J.-Y. Lee, and I.~S. Kweon, ``Cbam: Convolutional block
  attention module,'' in \emph{ECCV}, 2018, pp. 3--19.

\bibitem{attention2017}
A.~Vaswani, N.~Shazeer, N.~Parmar, J.~Uszkoreit, L.~Jones, A.~N. Gomez,
  L.~Kaiser, and I.~Polosukhin, ``Attention is all you need,'' in
  \emph{NeurIPS}, 2017, p. 6000–6010.

\bibitem{impactpe23}
A.~Kazemnejad, I.~Padhi, K.~N. Ramamurthy, P.~Das, and S.~Reddy, ``The impact
  of positional encoding on length generalization in transformers,'' in
  \emph{NeurIPS}, 2023, pp. 24\,892--24\,928.

\bibitem{roformer24}
J.~Su, M.~Ahmed, Y.~Lu, S.~Pan, W.~Bo, and Y.~Liu, ``Roformer: Enhanced
  transformer with rotary position embedding,'' \emph{Neurocomputing}, vol.
  568, no.~C, Feb. 2024.

\bibitem{llama23}
H.~Touvron, T.~Lavril, G.~Izacard, X.~Martinet, M.-A. Lachaux, T.~Lacroix,
  B.~Rozière, N.~Goyal, E.~Hambro, F.~Azhar, A.~Rodriguez, A.~Joulin,
  E.~Grave, and G.~Lample, ``Llama: Open and efficient foundation language
  models,'' in \emph{arXiv}, 2023.

\bibitem{bloom23}
T.~L. Scao, A.~Fan, C.~Akiki, E.~Pavlick, S.~Ilić, D.~Hesslow, and e.~a.
  Roman~Castagné, ``Bloom: A 176b-parameter open-access multilingual language
  model,'' in \emph{arXiv}, 2023.

\bibitem{upernet2018}
T.~Xiao, Y.~Liu, B.~Zhou, Y.~Jiang, and J.~Sun, ``Unified perceptual parsing
  for scene understanding,'' in \emph{ECCV}, 2018, pp. 418--434.

\bibitem{layerscale21}
H.~Touvron, M.~Cord, A.~Sablayrolles, G.~Synnaeve, and H.~Jégou, ``Going
  deeper with image transformers,'' in \emph{ICCV}, 2021, pp. 32--42.

\bibitem{polarmix_2022}
A.~Xiao, J.~Huang, D.~Guan, K.~Cui, S.~Lu, and L.~Shao, ``Polarmix: A general
  data augmentation technique for lidar point clouds,'' in \emph{NeurIPS},
  vol.~35, 2022, pp. 11\,035--11\,048.

\bibitem{lasermix_2023}
L.~Kong, J.~Ren, L.~Pan, and Z.~Liu, ``Lasermix for semi-supervised lidar
  semantic segmentation,'' in \emph{CVPR}, 2023, pp. 21\,705--21\,715.

\bibitem{learnpe17}
J.~Gehring, M.~Auli, D.~Grangier, D.~Yarats, and Y.~N. Dauphin, ``Convolutional
  sequence to sequence learning,'' in \emph{ICML}, 2017, p. 1243–1252.

\bibitem{squeezesegv2}
B.~Wu, X.~Zhou, S.~Zhao, X.~Yue, and K.~Keutzer, ``Squeezesegv2: Improved model
  structure and unsupervised domain adaptation for road-object segmentation
  from a lidar point cloud,'' in \emph{ICRA}, 2019, pp. 4376--4382.

\bibitem{squeezesegv3_2020}
C.~Xu, B.~Wu, Z.~Wang, W.~Zhan, P.~Vajda, K.~Keutzer, and M.~Tomizuka,
  ``Squeezesegv3: Spatially-adaptive convolution for efficient point-cloud
  segmentation,'' in \emph{ECCV}, 2020, pp. 1--19.

\bibitem{rangeformer_2023}
L.~Kong, Y.~Liu, R.~Chen, Y.~Ma, X.~Zhu, Y.~Li, Y.~Hou, Y.~Qiao, and Z.~Liu,
  ``Rethinking range view representation for lidar segmentation,'' in
  \emph{ICCV}, 2023, pp. 228--240.

\bibitem{empir24}
J.~Sun, C.~Qing, X.~Xu, L.~Kong, Y.~Liu, L.~Li, C.~Zhu, J.~Zhang, Z.~Xiao,
  R.~Chen, T.~Wang, W.~Zhang, and K.~Chen, ``An empirical study of training
  state-of-the-art lidar segmentation models,'' in \emph{arXiv}, 2024.

\bibitem{minet_2021}
S.~Li, X.~Chen, Y.~Liu, D.~Dai, C.~Stachniss, and J.~Gall, ``Multi-scale
  interaction for real-time lidar data segmentation on an embedded platform,''
  \emph{IEEE RAL}, vol.~7, no.~2, pp. 738--745, 2021.

\bibitem{cylindrical3d2021}
X.~Zhu, H.~Zhou, T.~Wang, F.~Hong, Y.~Ma, W.~Li, H.~Li, and D.~Lin,
  ``Cylindrical and asymmetrical 3d convolution networks for lidar
  segmentation,'' in \emph{CVPR}, 2021, pp. 9934--9943.

\bibitem{openpcseg2023}
Y.~Liu, R.~Chen, X.~Li, L.~Kong, Y.~Yang, Z.~Xia, Y.~Bai, X.~Zhu, Y.~Ma, Y.~Li,
  Y.~Qiao, and Y.~Hou, ``Uniseg: A unified multi-modal lidar segmentation
  network and the openpcseg codebase,'' in \emph{ICCV}, 2023, pp.
  21\,662--21\,673.

\bibitem{cityscapes16}
M.~Cordts, M.~Omran, S.~Ramos, T.~Rehfeld, M.~Enzweiler, R.~Benenson,
  U.~Franke, S.~Roth, and B.~Schiele, ``The cityscapes dataset for semantic
  urban scene understanding,'' in \emph{CVPR}, 2016, pp. 3213--3223.

\end{thebibliography}

\vfill

\clearpage

\onecolumn

%%%%%%%%%%%%%%%%%%%%%%%%%%%%%%%%%%%%%%%%%%%%%%%%%%%%%%%%%%%%%%%%%%%%%%%%%%%%%%%%
\section*{Supplementary Materials}

The supplementary materials are summarized as follows:
\begin{itemize}
	\item How to visualize the color and gray images as points (see Fig.~\ref{fig:color_gray_range_image})?
	
	\item How to prepare a range image with scan unfolding++~\cite{filling_missing2024}?
	
	\item Comparisons among feature maps of Fast FMVNet V2 and Fast FMVNet V3.
	
	\item Explanations of why spatial attention modules cannot work well for the range image-based models (see Table~\ref{tab:dam_attention_modules}).
	
	\item More comparisons among different range image-based models on the SemanticKITTI~\cite{semantickitti_2019_behley} validation dataset.
	
	\item Qualitative segmentation results of Fast FMVNet V3 on the SemanticKITTI~\cite{semantickitti_2019_behley}, nuScenes~\cite{nuscenes_panoptic}, and SemanticPOSS~\cite{semanticposs_2020} datasets. 
	 
\end{itemize}

\section{How to visualize the color and gray images as points?}
When visualizing the color image as points in Fig.~\ref{fig:color_gray_range_image}(a), we directly use the R, G, and B values as the $(x, y, z)$ coordinates. Besides, the RGB values serve as the colors. 

\begin{figure}[htp]
	\centering
	\includegraphics[width=0.5\columnwidth]{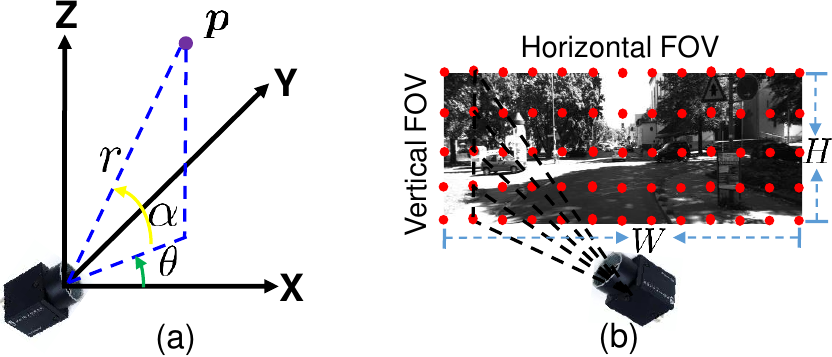}
	\caption{(a) A spherical model. (b) We see the gray image as the range image when visualizing the gray image as points.}
	\label{fig:spherical_model}  
\end{figure}

When visualizing the gray image as points in Fig.~\ref{fig:color_gray_range_image}(b), we see the gray image as the range image and adopt the spherical model to calculate the $(x, y, z)$ coordinates. The spherical model is expressed in Fig.~\ref{fig:spherical_model} and the following Eq.~(\ref{eq:spherical_model}),
\begin{equation}
	\begin{cases} 
		x = r \times \text{cos}(\alpha) \times \text{cos}(\theta), \\
		y = r \times \text{cos}(\alpha) \times \text{sin}(\theta), \\
		z = r \times \text{sin}(\alpha),
	\end{cases}
	\label{eq:spherical_model}
\end{equation}
where $r$ is a pixel value, and $\alpha$ and $\theta$ are the vertical and horizontal angles of the pixel. The $\alpha$ and $\theta$ can be obtained by the Eq.~(\ref{eq:alpha_theta_gray_img}),
\begin{equation}
	\begin{cases} 
	\alpha_{row} = (hvfov - lvfov) / H * row - lvfov, \\ 
	\theta_{col} = (hhfov - lhfov) / W * col - lhfov,
	\end{cases}
	\label{eq:alpha_theta_gray_img}
\end{equation}
where $hvfov$ and $lvfov$ are the highest and lowest values in the vertical field of view. $hhfov$ and $lhfov$ are the highest and lowest values in the horizontal field of view. $row$ and $col$ indicate the row and column of the pixel. $H$ and $W$ are the width and height of the image. Here, we set $hvfov$ and $lvfov$ to $-90^{\circ}$ and $90^{\circ}$, respectively. Also, $hhfov$ and $lhfov$ are set to $-15^{\circ}$ and $15^{\circ}$, respectively.

\section{How to prepare a range image with scan unfolding++?}
\begin{figure}[htp]
	\centering
	\includegraphics[width=0.4\columnwidth]{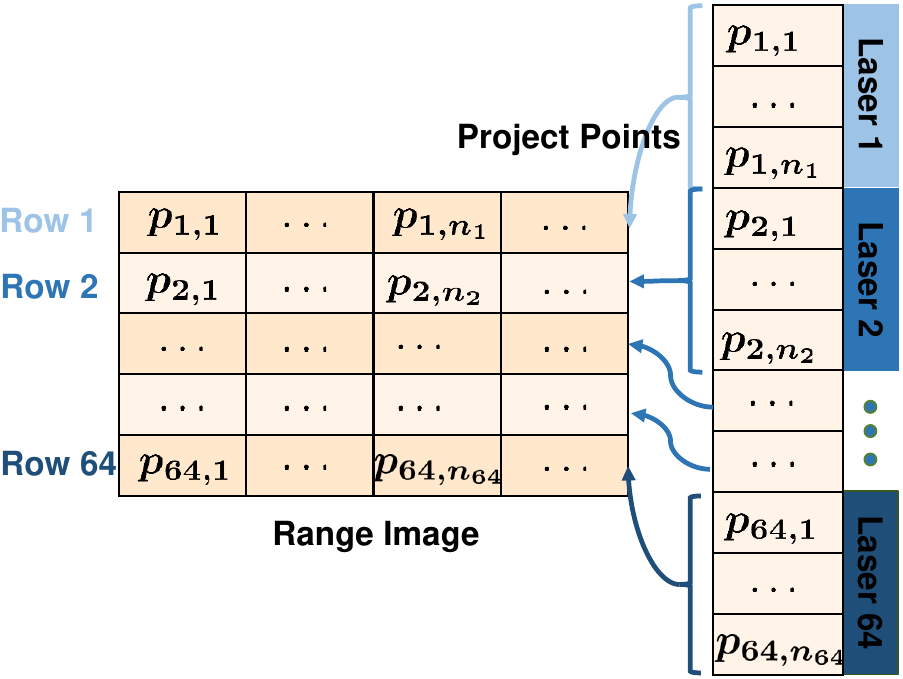}
	\caption{Preparation of the range image.}
	\label{fig:project_points_range_image}
\end{figure}

The range image prepared by scan unfolding++~\cite{filling_missing2024} keeps the shapes and patterns of objects. The method is described in Alg.~\ref{alg:scan_unfolding++} and Fig.~\ref{fig:project_points_range_image}. In Alg.~\ref{alg:scan_unfolding++}, Lines 1$\sim$4 aim to compute the horizontal angles of all points, and the angles are converted into positive values theoretically from $0^{\circ}$ to $360^{\circ}$. Line 5 transforms all horizontal angles to the corresponding horizontal coordinates $\boldsymbol{u}$ in the range image. Line 6 shows that the vertical coordinates $\boldsymbol{v}$ are identical to the ring numbers $\boldsymbol{d}$. With Alg.~\ref{alg:scan_unfolding++}, we can build a look-up table (LUT) connecting the points and the $\left(\boldsymbol{u}, \boldsymbol{v}\right)$ coordinates in the range image. Then, we project the points onto the range image. Specifically, as shown in Fig.~\ref{fig:project_points_range_image}, the points from the same laser are sequentially projected onto the corresponding row of the range image.

\begin{algorithm}[htp]
	\caption{Scan Unfolding++~\cite{filling_missing2024}.}
	\label{alg:scan_unfolding++}
	\small
	\begin{algorithmic}[1]
		\Require $N$ points in a point cloud expressed by $\boldsymbol{P}$; The ring numbers of all points $\boldsymbol{d}$; The width of the range image $W$.
		\Ensure $\left(\boldsymbol{u}, \boldsymbol{v}\right)$ coordinates for all points in the range image.
		
		\vspace{2ex}
		
		\State $\boldsymbol{x} = \boldsymbol{P}[:, 0]$, $\boldsymbol{y} = \boldsymbol{P}[:, 1]$. 
		
		\State $\boldsymbol{\theta} = \text{arctan}(\boldsymbol{y} / \boldsymbol{x}) \times 180 / \pi$.
		
		\State $\boldsymbol{m} = \boldsymbol{\theta} < 0$.
		
		\State $\boldsymbol{\theta}[\boldsymbol{m}] = \boldsymbol{\theta}[\boldsymbol{m}] + 360$.
		
		\State $\boldsymbol{u} = \boldsymbol{\theta} / 360 \times W$.
		
		\State $\boldsymbol{v} = \boldsymbol{d}$.
		
	\end{algorithmic}
\end{algorithm}

\section{Comparisons among feature maps of Fast FMVNet V2 and Fast FMVNet V3}
We here compare the feature maps output from Fast FMVNet V2~\cite{pdm2024} and our Fast FMVNet V3 to show the effectiveness of the proposed depth-aware module (DAM).

\begin{figure}[htp]
	\centering
	\includegraphics[width=1.0\columnwidth]{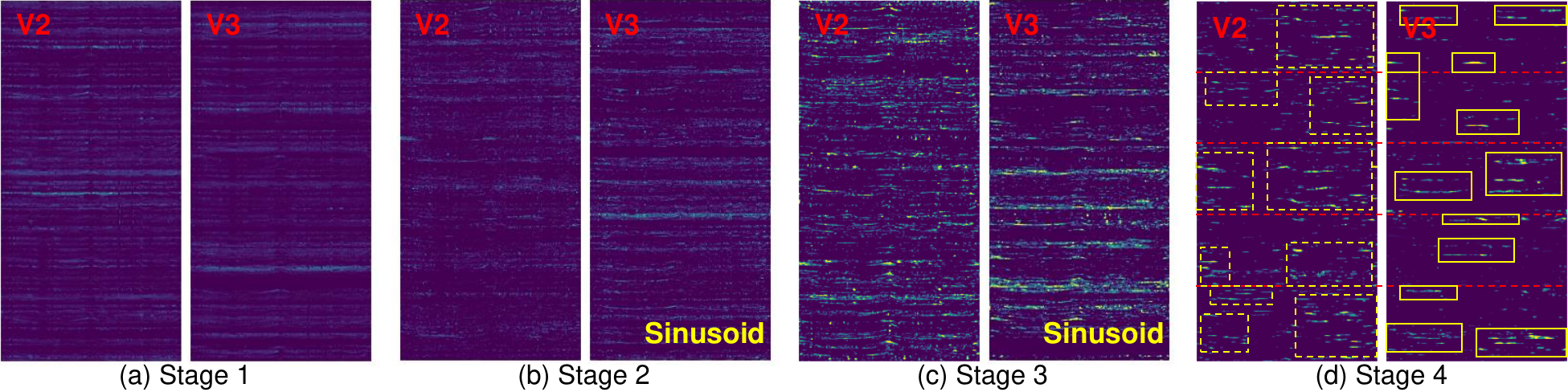}
	\caption{Feature maps from four Stages of Fast FMVNet V2~\cite{pdm2024} and our Fast FMVNet V3. ``V2" and ``V3" mean Fast FMVNet V2 and Fast FMVNet V3, respectively. For each picture, we sequentially stack the first 64 activation maps along the vertical direction. We see ``sinusoid" appearance along the vertical direction in the Fast FMVNet V3 feature maps in (b) and (c). Besides, in (d), high response regions in Fast FMVNet V3 are sparser than these in Fast FMVNet V2.}
	\label{fig:activation_maps}
\end{figure}

The qualitative comparison results are provided in Fig.~\ref{fig:activation_maps}. The feature maps are obtained after the DAM in Fast FMVNet V3 (see \textcolor{red}{D}ConvNeXt Block in Fig.~\ref{fig:fast_fmvnet_v3}). Correspondingly, we get the feature maps after GELU in Fast FMVNet V2 (see ConvNeXt Block in Fig.~\ref{fig:fast_fmvnet_v3}). In Figs.~\ref{fig:activation_maps}(b) and (c), we see ``sinusoid" appearance along the vertical direction in the Fast FMVNet V3 feature maps, suggesting that the position information across channels is learned by the model and hence the ordered depth information in the range image is perceived. In Fig.~\ref{fig:activation_maps}(d), high response regions in the Fast FMVNet V3 feature maps are sparser than those in the Fast FMVNet V2 feature maps. If we move all yellow rectangles to one patch (\textit{i.e.}, emphasized by the red dash rectangles), the overlapping area of the yellow rectangles in V3 is smaller than that in V2, suggesting that the introduced DAM enables the model to focus on different regions at different feature maps. This also indicates that DAM might be a good regularizer and promote the generalization of Fast FMVNet V3. For example, Fast FMVNet V3 obtains 68.84\% on the SemanticKITTI validation dataset but gets 69.6\% (+0.76) on the testing dataset. By contrast, Fast FMVNet V2 achieves 68.55\% and 68.1\% (-0.45) on the validation and testing datasets, respectively. 

\begin{table*}[htp]
	\caption{Comparison results between Fast FMVNet V2~\cite{pdm2024} and Fast FMVNet V3 regarding the cosine distance among feature maps.}
	\label{tab:cosine_similarity}
	\centering
	\scalebox{1.0}{
		\begin{tabular}{l|l|l|l|l}
			Model           & Stage 1 $\uparrow$  & Stage 2 $\uparrow$  & Stage 3 $\uparrow$  & Stage 4 $\uparrow$ \\ \hline
			Fast FMVNet V2  & 0.476               & 0.477               & 0.461               & 0.480          \\
			Fast FMVNet V3  & 0.478 (\textcolor{blue}{+0.002})      & 0.478 (\textcolor{blue}{+0.001})      & 0.466 (\textcolor{blue}{+0.005})      & 0.486 (\textcolor{blue}{+0.006}) \\
		\end{tabular}}
\end{table*}

We further use cosine distance~\cite{convnextv223} among feature channels to complement the above qualitative comparisons. The cosine distance is expressed in the following Eq.~(\ref{eq:cosine_distance}),
\begin{equation}
	dist = \frac{1}{C^2}\sum_{i=0}^{C}\sum_{j=0}^{C} \frac{1-\text{cos}(\boldsymbol{M}_{i,:,:}, \boldsymbol{M}_{j,:,:})}{2},
	\label{eq:cosine_distance}
\end{equation}
where $\boldsymbol{M} \in \mathbb{R}^{C \times H \times W}$ indicates the feature maps with channels $C$, height $H$, and width $W$. A higher distance value means more diverse features learned by the model. In experiments, we randomly choose 100 point clouds on the SemanticKITTI validation dataset to compute the average cosine distance values for four Stages. The results are reported in Table~\ref{tab:cosine_similarity}. We see that with DAM, Fast FMVNet V3 learns more diverse features than Fast FMVNet V2, especially in Stages 3 and 4. The quantitative results also explain why the high response regions in the Fast FMVNet V3 feature maps in Fig.~\ref{fig:activation_maps}(d) are sparser than these in the Fast FMVNet V2 feature maps, and DAM improves the generalization of Fast FMVNet V3.

\section{Explanations of why spatial attention modules cannot work well for the range image-based models}
In Table~\ref{tab:dam_attention_modules}, the spatial attention modules, such as CBAM-Spatial Attention~\cite{cbam18}, Pooling~\cite{metaformer22}, Multi-Scale Convolutional Attention~\cite{segnext22}, and Focal Modulation~\cite{focalnet22}, cannot improve the performance of Fast FMVNet V3, probably because of the difference between the color image and the range image. 
\begin{figure}[htp]
	\centering
	\includegraphics[width=0.45\columnwidth]{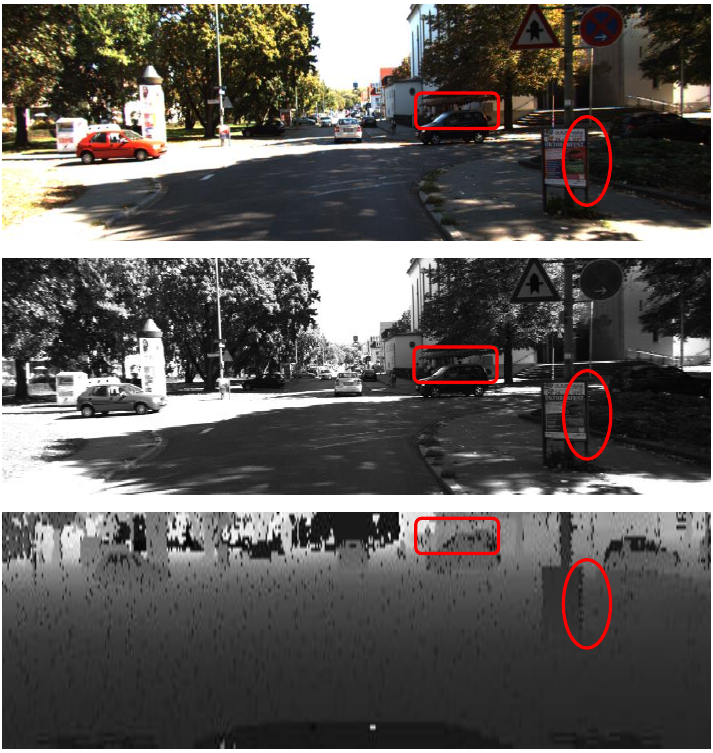}
	\caption{Color, gray, and range images. It is difficult to see the boundaries between the car roof and the building and between the traffic sign and the terrain in the color and gray images. By contrast, we can easily find the corresponding boundaries in the range image.}
	\label{fig:color_gray_range_image_comparison}
\end{figure}
Specifically, all images are expressed by matrices. Deep learning models recognize objects by dynamically adjusting the parameters to fit pixel values to highlight the shapes and other patterns of the objects. In the color and gray images, the boundary pixel values between different objects are very close, such as the boundaries between the car roof and the building and between the traffic sign and the terrain in Fig.~\ref{fig:color_gray_range_image_comparison}, bringing difficulty for the models to accurately segment these objects. Therefore, the spatial attention modules are required to generate weights to explicitly emphasize the object shapes and patterns further. By contrast, in the range image, the boundary depth values between different objects are not close (see the range image in Fig.~\ref{fig:color_gray_range_image_comparison}). Therefore, the spatial attention modules for the range image-based models are unnecessary.

\section{More comparisons among different range image-based models on the SemanticKITTI validation dataset}
\begin{figure}[htp]
	\centering
	\includegraphics[width=0.4\columnwidth]{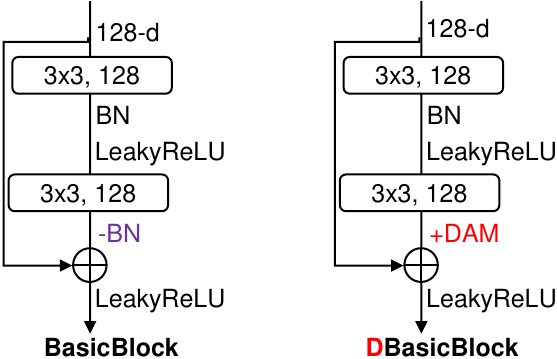}
	\caption{In FIDNet~\cite{fidnet_2021} and CENet~\cite{cenet_2022}, the second batch normalization (BN) in BasicBlock is replaced with the introduced depth-aware module (\textcolor{red}{DAM}) to build the depth-aware BasicBlock (\textcolor{red}{D}BasicBlock).}
	\label{fig:fidnet_cenet}
\end{figure}

We here provide more comparison results among various range image-based models on the SemanticKITTI~\cite{semantickitti_2019_behley} validation dataset. To validate the effectiveness of the proposed depth-aware module (DAM), we integrate it into FIDNet~\cite{fidnet_2021,pdm2024} and CENet~\cite{cenet_2022,pdm2024}. Specifically, as shown in Fig.~\ref{fig:fidnet_cenet}, we replace the second batch normalization in BasicBlock with DAM to build the depth-aware BasicBlock (\textcolor{red}{D}BasicBlock). Also, in the FIDNet and CENet architectures, only the last BasicBlock in each stage is replaced with \textcolor{red}{D}BasicBlock. The upgraded models, FIDNet+DAM and CENet+DAM, are trained on the SemanticKITTI training dataset, and results are reported on the validation dataset. The comparison results are provided in Table~\ref{tab:64x2048_kitti_val_results}. We see that FIDNet+DAM and CENet+DAM consistently achieve better performance than their counterparts (\textit{i.e.}, 67.3\% (+1.1) \textit{vs.} 66.2\% and 66.9\% (+0.6) \textit{vs.} 66.3\%, respectively). The performance gains validate the effectiveness of the proposed DAM.

\begin{table*}[htp]
	\caption{Comparison results among range image-based models on the SemanticKITTI~\cite{semantickitti_2019_behley} val set regarding IoU and mIoU scores (\%). ``$\ast$": the models pre-trained on the Cityscapes~\cite{cityscapes16} dataset. Note that \textbf{none of test-time augmentation and ensemble tricks} are applied to boost model performance. \textbf{Bold} font: the best results. \underline{Underline}: the second-best results.}
	\label{tab:64x2048_kitti_val_results}
	\centering
	\scalebox{0.83}{
		\begin{tabular}{l|c|c|c|c|c|c|c|c|c|c|c|c|c|c|c|c|c|c|c|c}
			Models	 & mIoU &\rotatebox{90}{Car} &\rotatebox{90}{Bicycle} &\rotatebox{90}{Motorcycle} &\rotatebox{90}{Truck} &\rotatebox{90}{Other-vehicle} &\rotatebox{90}{Person} &\rotatebox{90}{Bicyclist} &\rotatebox{90}{Motorcyclist} &\rotatebox{90}{Road} &\rotatebox{90}{Parking} &\rotatebox{90}{Sidewalk} &\rotatebox{90}{Other-ground} &\rotatebox{90}{Building} &\rotatebox{90}{Fence} &\rotatebox{90}{Vegetation} &\rotatebox{90}{Trunk} &\rotatebox{90}{Terrain} &\rotatebox{90}{Pole} &\rotatebox{90}{Traffic-sign}  \\ \hline 
			
			RangeNet53~\cite{pdm2024}&64.8 &95.7 &50.4 &73.4 &75.3 &50.4 &77.5 &87.7  &0.0 &\underline{95.9} &51.1 &83.4 &6.9 &90.2 &63.4 &86.4 &64.5 &72.5 &56.7 &50.7\\ 
			
			FIDNet~\cite{pdm2024}&66.2 &95.0 &\textbf{53.7} &72.9 &88.1 &59.0 &80.0 &90.3 &0.0  &94.9 &42.1 &82.6 &10.0 &90.0 &59.3 &86.9 &67.3 &72.7 &61.9 &51.6 \\ 
			\textcolor{blue}{+DAM}&\textcolor{blue}{67.3} &94.8 &52.5 &70.5 &88.2 &63.7 &81.5 &89.9 &0.0  &95.1 &44.3 &82.9 &\textbf{25.5} &90.2 &58.4 &86.9 &67.6 &72.7 &61.9 &\underline{52.9} \\ 
			 
			CENet~\cite{pdm2024} &66.3 &94.1 &50.5 &72.0 &83.2 &56.8 &81.8 &91.5 &0.0  &94.9 &41.8 &81.8 &13.5 &90.7 &59.7 &87.7 &70.4 &75.4 &60.4 &\textbf{54.0} \\ 
			\textcolor{blue}{+DAM}&\textcolor{blue}{66.9} &94.7 &47.2 &65.1 &\underline{91.4} &56.8 &81.8 &90.2 &\underline{0.1} &95.3 &43.6 &82.9 &25.1 &90.4 &57.1 &\textbf{88.4} &\textbf{71.0} &\textbf{76.4} &61.6 &52.3 \\ 
			
			Fast FMVNet        &67.4 &96.1 &50.3 &74.0 &88.6 &\underline{67.4} &\underline{82.2} &91.1 &0.0 &95.5 &49.2 &83.8 &9.1 &90.6 &63.0 &86.1 &70.4 &70.1 &63.8 &50.2 \\ 
			Fast FMVNet$^\ast$ &67.9 &95.3 &52.2 &\underline{76.9} &\textbf{91.6} &52.0 &80.8 &\underline{91.7} &\textbf{0.1} &95.8 &\textbf{60.7} &\underline{84.5} &13.9 &\underline{91.2} &65.1 &86.4 &70.2 &70.9 &61.6 &49.3 \\ 
			
			Fast FMVNet V2        &68.5 &\underline{96.6} &46.5 &76.0 &87.6 &66.5 &81.4 &91.0 &0.0 &95.8 &\underline{57.4} &84.1 &16.6 &90.8 &\underline{65.6} &\underline{87.9} &69.3 &\underline{75.6} &61.5 &52.0 \\
			Fast FMVNet V2$^\ast$ &\underline{68.6} &\textbf{96.8} &51.0 &\textbf{80.9} &85.8 &\textbf{68.5} &\textbf{82.4} &90.8 &0.0 &\textbf{96.2} &48.1 &\textbf{84.7} &9.5  &\textbf{92.2} &\textbf{68.8} &87.4 &\underline{70.8} &72.6 &\textbf{64.8} &52.2 \\ 
			
			\textcolor{blue}{Fast FMVNet V3} &\textcolor{blue}{\textbf{68.8}} &96.4 &\underline{53.0} &73.9 &90.2 &63.6 &81.4 &\textbf{91.9} &0.0 &95.7 &53.7 &84.3 &\underline{25.3} &90.9 &64.9 &86.4 &70.6 &70.9 &\underline{63.8} &51.0 \\	\end{tabular}}
\end{table*}

\section{Qualitative segmentation results of Fast FMVNet V3}
\begin{figure}[htp]
	\centering
	\includegraphics[width=1.0\columnwidth]{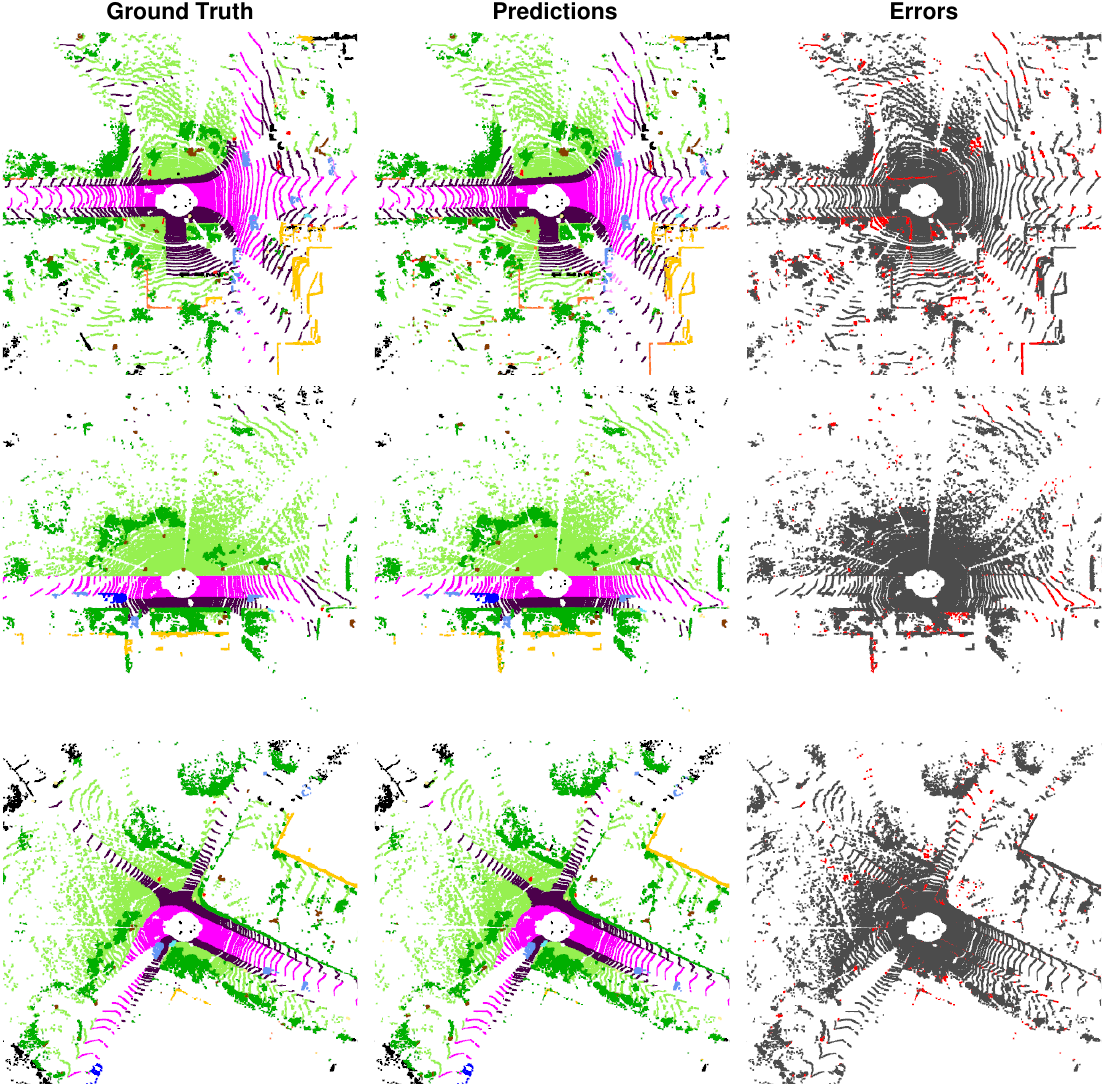}
	\caption{Qualitative segmentation results on the SemanticKITTI~\cite{semantickitti_2019_behley} validation dataset. In the last column, the \textcolor{red}{incorrect} predictions are emphasized by the \textcolor{red}{red} color.}
	\label{fig:kitti_qualitative}
\end{figure}

\begin{figure}[htp]
	\centering
	\includegraphics[width=1.0\columnwidth]{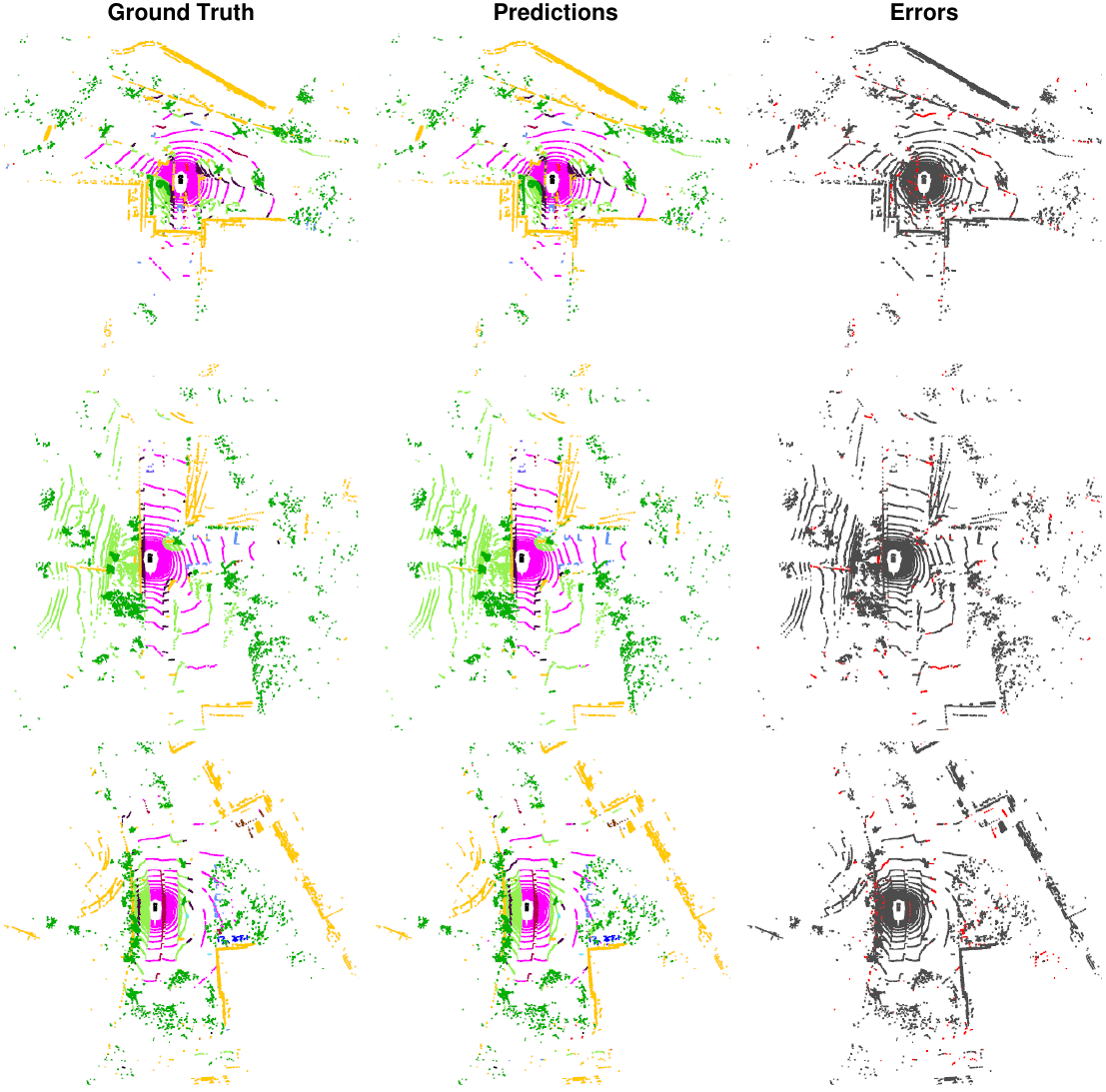}
	\caption{Qualitative segmentation results on the nuScenes~\cite{nuscenes_panoptic} validation dataset. In the last column, the \textcolor{red}{incorrect} predictions are emphasized by the \textcolor{red}{red} color.}
	\label{fig:nuscenes_qualitative}
\end{figure}

\begin{figure}[htp]
	\centering
	\includegraphics[width=1.0\columnwidth]{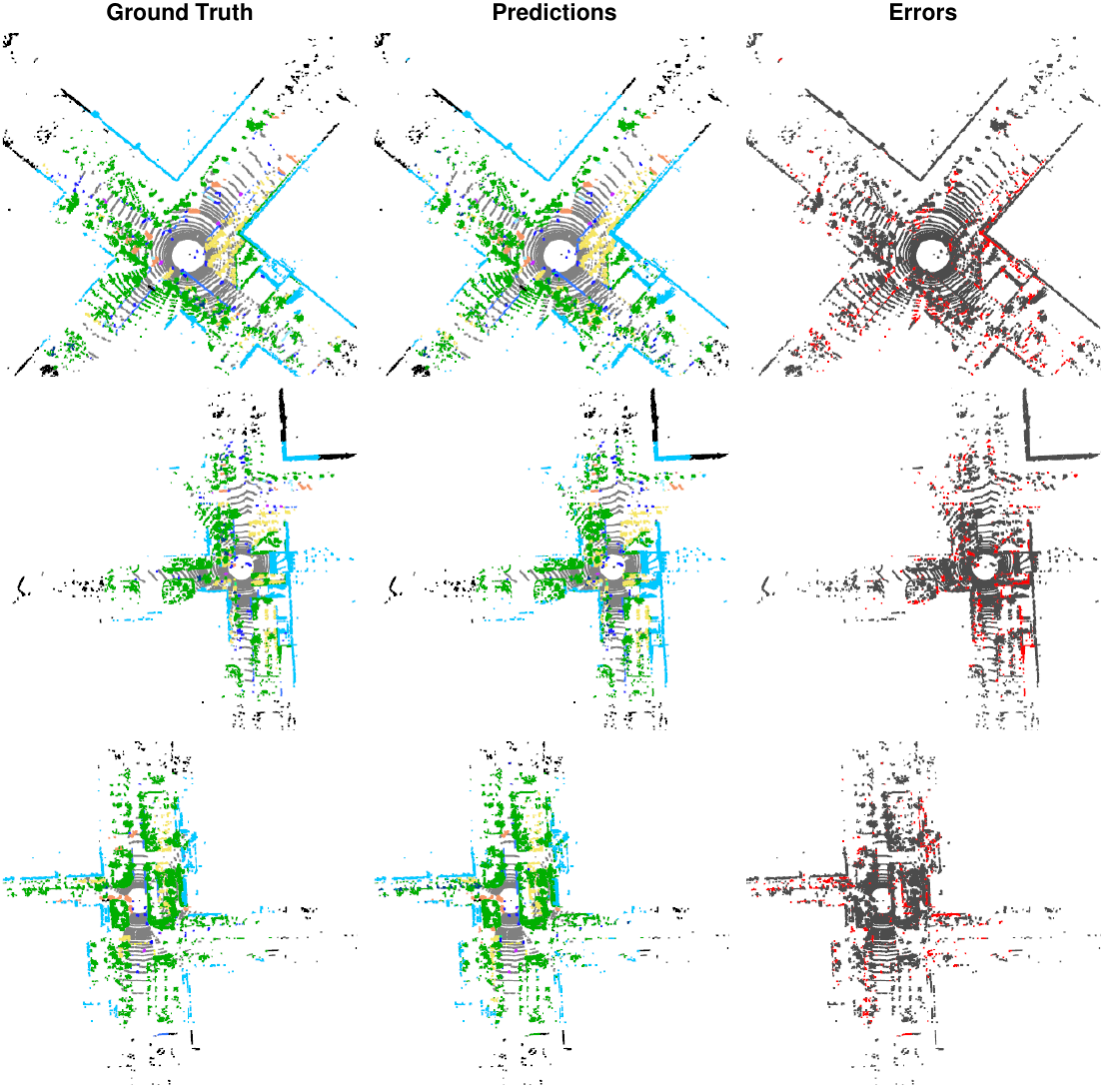}
	\caption{Qualitative segmentation results on the SemanticPOSS~\cite{semanticposs_2020} test dataset. In the last column, the \textcolor{red}{incorrect} predictions are emphasized by the \textcolor{red}{red} color.}
	\label{fig:poss_qualitative}
\end{figure}

We here show qualitative segmentation results of Fast FMVNet V3 on the SemanticKITTI~\cite{semantickitti_2019_behley}, nuScenes~\cite{nuscenes_panoptic}, and SemanticPOSS~\cite{semanticposs_2020} datasets in Figs.~\ref{fig:kitti_qualitative},~\ref{fig:nuscenes_qualitative}, and~\ref{fig:poss_qualitative}.

\end{document}